\documentclass[preprint,12pt]{elsarticle}
\ifdefined\pdfminorversion
\fi
\usepackage[T1]{fontenc}
\usepackage{lmodern}
\usepackage[a4paper,margin=1in]{geometry}
\usepackage{microtype}
\usepackage{amsmath}
\usepackage{graphicx}
\usepackage{booktabs}
\usepackage{array}
\usepackage{tabularx}
\usepackage{longtable}
\usepackage{pdflscape}
\usepackage{enumitem}
\usepackage[table,dvipsnames]{xcolor}
\usepackage{caption}
\usepackage{float}
\usepackage[bookmarks=false]{hyperref}
\usepackage{xurl}
\usepackage{seqsplit}
\providecommand{\parencite}{\citep}
\providecommand{\textcite}{\citet}
\usepackage{multirow}

\journal{Neurocomputing}

\hypersetup{
  colorlinks=true,
  linkcolor=blue!45!black,
  citecolor=blue!45!black,
  urlcolor=blue!45!black,
  pdftitle={Uncertainty-Aware Decision Making in Multimodal Large Language Models},
  pdfauthor={Anonymous Author(s)}
}

\setlist[itemize]{topsep=2pt,itemsep=2pt,parsep=0pt,leftmargin=1.6em}
\setlist[enumerate]{topsep=2pt,itemsep=2pt,parsep=0pt,leftmargin=1.7em}

\newcolumntype{Y}{>{\raggedright\arraybackslash}X}
\newcolumntype{P}[1]{>{\raggedright\arraybackslash}p{#1}}
\graphicspath{{./}}

\begin{document}

\begin{frontmatter}

\title{Uncertainty-Aware Decision Making in Multimodal Large Language Models: A Survey of Sources, Signals, Calibration, and Actions}

\author[label1]{Abderrahmene Boudiaf\corref{cor1}}
\ead{100058322@ku.ac.ae}
\author[label1]{Irfan Hussain}
\author[label1]{Sajid Javed}
\address[label1]{Khalifa University of Science and Technology, Abu Dhabi, United Arab Emirates}
\cortext[cor1]{Corresponding author}

\begin{abstract}
Multimodal large language models (MLLMs) increasingly answer questions whose correctness depends on visual, textual, temporal, acoustic, document, chart, or embodied evidence. Their failures are therefore not only linguistic. A fluent answer may conceal poor input quality, a perceptual error, weak grounding, conflict between modalities, unstable reasoning, distribution shift, or a question that is not answerable from the supplied evidence. This survey organizes the literature on uncertainty-aware MLLMs around a decision-centered framework: uncertainty sources give rise to observable signals, signals must be calibrated or controlled for risk, and calibrated uncertainty should determine the system action. We review work on token and logit uncertainty, semantic disagreement, perturbation instability, grounding and attribution scores, verbalized confidence, verifier and judge scores, conformal prediction, selective answering, abstention, clarification, retrieval, self-checking, and escalation. The central argument is that uncertainty should not be evaluated only as a confidence number; it should be evaluated by whether it improves behavior under insufficient, conflicting, shifted, or high-risk multimodal evidence. We position this survey against text-only uncertainty and abstention surveys, broad MLLM surveys, MLLM hallucination surveys, and safety-oriented reviews. We conclude with open problems in source-aware decomposition, action-aware benchmarks, calibration under shift, black-box uncertainty estimation, broader modality coverage, reproducible reporting, and human-centered uncertainty communication.
\end{abstract}


\end{frontmatter}

\section{Introduction}
\label{sec:introduction}

Multimodal large language models extend language generation with perceptual and structured inputs, including images, video, audio, charts, documents, and embodied observations. This expansion has made MLLMs useful for visual question answering, chart and document understanding, video reasoning, medical image interpretation, multimodal dialogue, and interactive agents \parencite{yin2023surveyMLLM,largeMultimodalEval2025}. It has also changed the reliability problem. A generated answer may be linguistically fluent while the underlying evidence is blurred, occluded, incomplete, conflicting, outside the model's training distribution, or simply insufficient to answer the question.

This distinction matters because uncertainty in MLLMs is not a single phenomenon. Some uncertainty begins in the input: the image may be low quality, the video may omit the relevant moment, or the audio may be noisy. Some uncertainty arises during perception and grounding: the model may recognize the wrong object, attend to the wrong region, or produce a response that is only weakly supported by the visual evidence. Other uncertainty appears later, when the language component fills missing evidence with a prior, when the reasoning path becomes unstable, or when the deployment domain differs from the calibration setting. In the most direct case, the question is not answerable from the supplied multimodal evidence at all.

Existing text-only work on uncertainty quantification, confidence calibration, semantic uncertainty, selective prediction, and abstention provides important foundations \parencite{shorinwa2025surveyuqllm,liu2025uqcalibrationsurvey,xia2025surveyuqllm,wen2025knowyourlimits}. These foundations remain relevant, but they are not sufficient for multimodal systems. A text-only confidence score does not reveal whether the evidence was visible, whether the visual and textual signals agreed, whether the answer was grounded in the correct region, or whether the model should have asked for a clearer input. For MLLMs, uncertainty must be connected to evidence quality and to the action that follows from that evidence.

The central thesis of this survey is therefore decision-centered: uncertainty is useful when it changes what the system does. A system that reports a confidence value but answers every question in the same manner is not yet uncertainty-aware in the deployment sense. A more reliable MLLM should answer when evidence is sufficient, hedge when confidence is limited, abstain or say ``I do not know'' when evidence is missing, ask for clarification when the user query is underspecified, request a better input when perception is weak, retrieve external evidence when the task requires knowledge beyond the input, self-check when reasoning is unstable, and escalate when the situation is too risky for autonomous handling.

\begin{figure}[H]
    \centering
    \includegraphics[width=\textwidth]{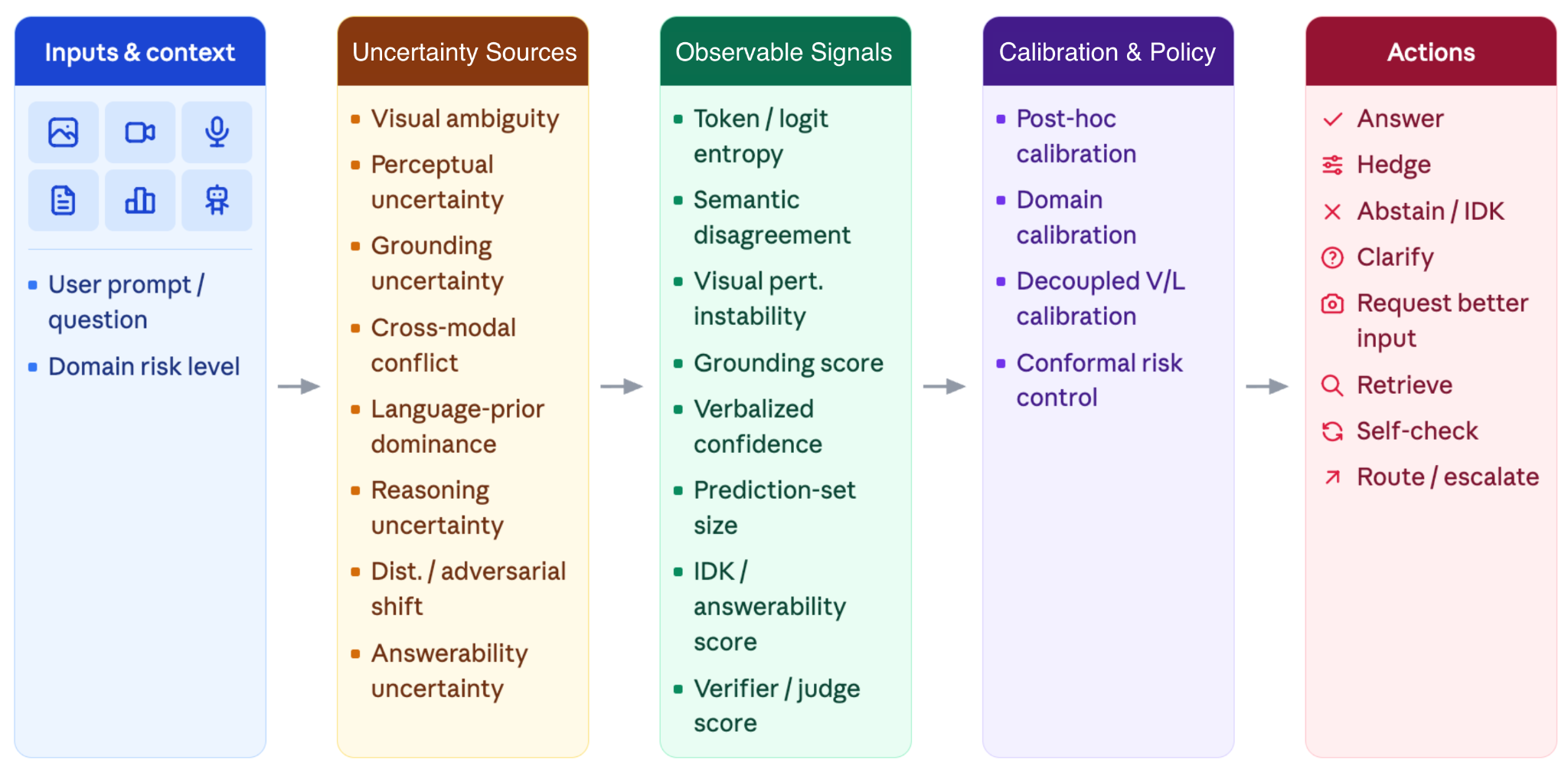}
    \caption{Source--signal--calibration--action framework for uncertainty-aware MLLMs. Multimodal inputs and domain context give rise to uncertainty sources. These sources are observed through model-internal, output-level, evidence-grounding, verbalized, verifier, or judge signals. Calibration and risk-control methods convert raw signals into decision-relevant estimates. The final policy selects an action: answer, hedge, abstain or say IDK, clarify, request better input, retrieve evidence, self-check, or route the case to a stronger model or human expert.}
    \label{fig:framework}
\end{figure}

Figure~\ref{fig:framework} gives the organizing framework for the paper and serves as the survey roadmap. Inputs and context give rise to uncertainty sources; sources produce observable signals; signals are converted into calibrated confidence or controlled risk; and the resulting estimate supports an action policy. This source--signal--calibration--action chain lets us discuss diverse methods without turning the survey into a list of isolated papers.

The evidence base around this framework spans calibration before and after multimodal training \parencite{chen2025unveiling}, reasoning-step confidence boundaries \parencite{he2025mmboundary}, decoupled visual and reasoning confidence \parencite{xiao2026vlcalibration}, training-free multimodal uncertainty estimation \parencite{lau2026umpire,bhattacharya2025festa}, selective visual question answering \parencite{eisenschlos2024selectively,wieczorek2026variational}, multimodal self-awareness and IDK behavior \parencite{wang2024mmsap,song2026visualidk}, conformal prediction and risk control \parencite{wang2025tron,tayebati2025cap,ye2025scpvlm,azad2026artmaybe,grc2026ocr}, modality-specific uncertainty benchmarks \parencite{wang2026vlmuqbench}, and refusal or answerability in visual, video, embodied, audio, and omni-modal settings \parencite{miyai2024upd,yoon2025videollmsrefuse,zhu2025mohobench,madhusudhan2026knowing,ortiz2026abstentionknobs,alnazi2026omdbench,abstaineqa2025robots,tran2026ravqa,aallm2026walking}. These studies are related, but they are scattered across calibration, hallucination detection, selective prediction, answerability, conformal risk control, and domain-specific trustworthiness. This survey brings them into one account centered on decision quality.

\paragraph{Contributions}
The survey makes four contributions. First, it introduces a source--signal--calibration--action taxonomy for uncertainty-aware MLLMs. Second, it synthesizes uncertainty sources that are specific to multimodal systems, including sensory ambiguity, perceptual error, grounding failure, cross-modal conflict, language-prior dominance, reasoning instability, answerability uncertainty, and shift. Third, it connects uncertainty estimation to response actions such as abstention, clarification, retrieval, self-checking, and escalation. Fourth, it reorganizes evaluation around behavior under uncertainty rather than confidence quality alone.

\paragraph{Scope}
The survey focuses on MLLMs, large vision-language models, and closely related VLM bridge work when the paper directly informs uncertainty estimation, calibration, risk control, answerability, abstention, or action selection. Text-only LLM uncertainty and abstention studies are used as background when they define method families, evaluation concepts, or useful baselines. General hallucination, safety, and broad MLLM surveys are discussed only when they clarify the boundary of the present survey.

\paragraph{Organization}
Section~\ref{sec:scope} defines the scope, positioning, and evidence base. Section~\ref{sec:sources} reviews sources of uncertainty in MLLMs. Section~\ref{sec:signals} organizes observable signals and estimation methods. Section~\ref{sec:calibration} discusses calibration and risk control. Section~\ref{sec:actions} connects uncertainty to response actions. Section~\ref{sec:evaluation} reviews evaluation protocols, benchmarks, and metrics. Section~\ref{sec:applications} discusses deployment settings. Section~\ref{sec:agenda} gives the research agenda. Sections~\ref{sec:survey-limitations} and~\ref{sec:conclusion} close the paper.

\section{Scope, Positioning, and Evidence Base}
\label{sec:scope}

A survey on uncertainty-aware MLLMs needs clear boundaries. We define uncertainty-aware decision making as the study of methods that estimate, express, calibrate, or control uncertainty in order to improve behavior under multimodal evidence. The definition includes uncertainty estimation, but it also includes the downstream policy: whether the system should answer, hedge, abstain, ask for clarification, request better input, retrieve evidence, verify its response, or escalate the case.

\subsection{Problem definition and terminology}

Let a user provide multimodal evidence $M$, a prompt or task $q$, and optional context $c$ such as domain, risk level, or retrieved information. An uncertainty-aware MLLM does not only produce an answer $y$. It also estimates a signal $s$, converts that signal into a calibrated confidence or risk estimate $r$, and selects an action $a$ through a policy $\pi$. Informally, the decision process is
\begin{equation}
    (a, y, r) = \pi\big(q, M, c, s(q,M,c)\big).
\end{equation}
The answer may be returned directly, qualified with a hedge, replaced by an IDK response, delayed until clarification or retrieval is performed, or routed to another system or human expert. This formulation is deliberately simple. Its purpose is to keep the survey focused on the relation between uncertainty and action rather than on a single mathematical definition of uncertainty.

We use \emph{uncertainty source} for the underlying reason a response may be unreliable. We use \emph{uncertainty signal} for an observable quantity or behavior that may reveal unreliability, such as entropy, logit margin, semantic disagreement, visual perturbation instability, grounding score, verbalized confidence, prediction-set size, answerability score, or verifier score. We use \emph{calibration} for the alignment between an estimated confidence or risk and empirical correctness or action cost. We use \emph{policy} for the rule that maps uncertainty to a response action.

\subsection{Inclusion criteria}

A paper is central to this survey when uncertainty is part of the method, benchmark, metric, or response policy. This includes MLLM and LVLM work on confidence estimation, calibration, selective answering, conformal prediction, hallucination or error detection when framed as uncertainty, answerability, IDK behavior, refusal, self-checking, retrieval decisions, and escalation. We also include bridge work from VLMs and text-only LLMs when it defines concepts or methods that are necessary for understanding MLLM uncertainty.

We do not attempt to survey every paper on multimodal hallucination, safety, robustness, or broad MLLM evaluation. Those literatures are adjacent and often relevant, but they become central here only when they connect evidence uncertainty to calibrated confidence, risk control, or action selection. This boundary prevents the survey from becoming a general MLLM reliability review.

\subsection{Position relative to adjacent surveys}

The closest survey families cover text-only uncertainty, text-only abstention, MLLM hallucination, MLLM safety, or broad MLLM evaluation. These areas are important, but they do not fully center the source-to-action chain for multimodal evidence. Table~\ref{tab:related-surveys} summarizes the positioning in categorical terms. This format is preferable to an arbitrary two-axis plot because the difference is conceptual rather than geometric.

\begin{table}[H]
\centering
\small
\caption{Positioning against adjacent survey families. The table separates related coverage from the gap addressed by this survey. Citations in the first four rows are representative anchors for adjacent survey families; the final row states the positioning of the present survey.}
\label{tab:related-surveys}
\begin{tabularx}{\linewidth}{@{}P{0.23\linewidth}YY@{}}
\toprule
\textbf{Survey family} & \textbf{Main coverage} & \textbf{Gap for this survey} \\
\midrule
Text-only LLM uncertainty & Uncertainty quantification, confidence calibration, semantic uncertainty, and selective prediction in language-only settings \parencite{shorinwa2025surveyuqllm,liu2025uqcalibrationsurvey,xia2025surveyuqllm,he2026surveyuncertaintysources}. & Limited treatment of visual evidence quality, grounding, cross-modal conflict, and multimodal answerability. \\
Text-only abstention and IDK & Knowledge boundaries, refusal, and selective answering in language-only settings \parencite{wen2025knowyourlimits}. & Abstention is not tied to visual sufficiency, image quality, or multimodal grounding. \\
MLLM hallucination and safety & Hallucination symptoms, attack and defense settings, benchmark construction, and safety risks \parencite{bai2024hallucinationsurvey,liu2024surveyhallucinationlvlm,liu2024mllmsafety,chen2026multimodalhallucinationsurvey}. & Often organized around failure detection or safety risk rather than uncertainty signals, calibration, and action policies. \\
Broad MLLM and evaluation surveys & Architectures, training data, capabilities, benchmarks, and application coverage \parencite{yin2023surveyMLLM,largeMultimodalEval2025}. & Too broad to compare uncertainty sources, signal access assumptions, calibration targets, and response actions in detail. \\
\textbf{This survey} & Source-aware uncertainty, observable signals, calibration, risk control, selective answering, abstention, clarification, retrieval, self-checking, escalation, and action-aware evaluation. & Integrates multimodal uncertainty estimation with decision policy. \\
\bottomrule
\end{tabularx}
\end{table}

\subsection{Evidence base and traceability}

The core evidence base contains 67 focused evidence rows drawn from MLLM and LVLM studies, VLM bridge work, domain-specific studies, and directly relevant surveys. The manuscript also uses additional background survey citations for positioning in Table~\ref{tab:related-surveys}; these background citations are not counted as focused evidence rows. The evidence matrix retains one additional non-counted bridge row for traceability. Search logs, evidence-tier accounting, extraction fields, and corpus-refresh details are retained in ~\ref{app:evidence}. This separation preserves traceability without interrupting the main synthesis. In the main synthesis tables, citations attached to row labels are representative anchors rather than exhaustive lists; explanatory, action, and caveat cells synthesize recurring patterns from the cited works and the surrounding subsections.

With the scope fixed, the rest of the paper follows the framework in Figure~\ref{fig:framework}. We first identify where uncertainty enters the multimodal pipeline, then ask how those sources can be observed, calibrated, converted into actions, and evaluated.

\section{Sources of Uncertainty in MLLMs}
\label{sec:sources}

The first question in the framework is where uncertainty comes from. For MLLMs, this question cannot be answered by looking only at the final answer or token distribution. The uncertainty may originate in the input, the perceptual encoder, the cross-modal alignment, the reasoning path, the deployment domain, or the answerability of the task. A source-aware taxonomy matters because different sources call for different actions. Blur may call for a better image. Missing evidence may call for abstention. Cross-modal conflict may call for self-checking or retrieval. A high-risk shifted domain may call for escalation.

\begin{figure}[H]
    \centering
    \includegraphics[width=\textwidth]{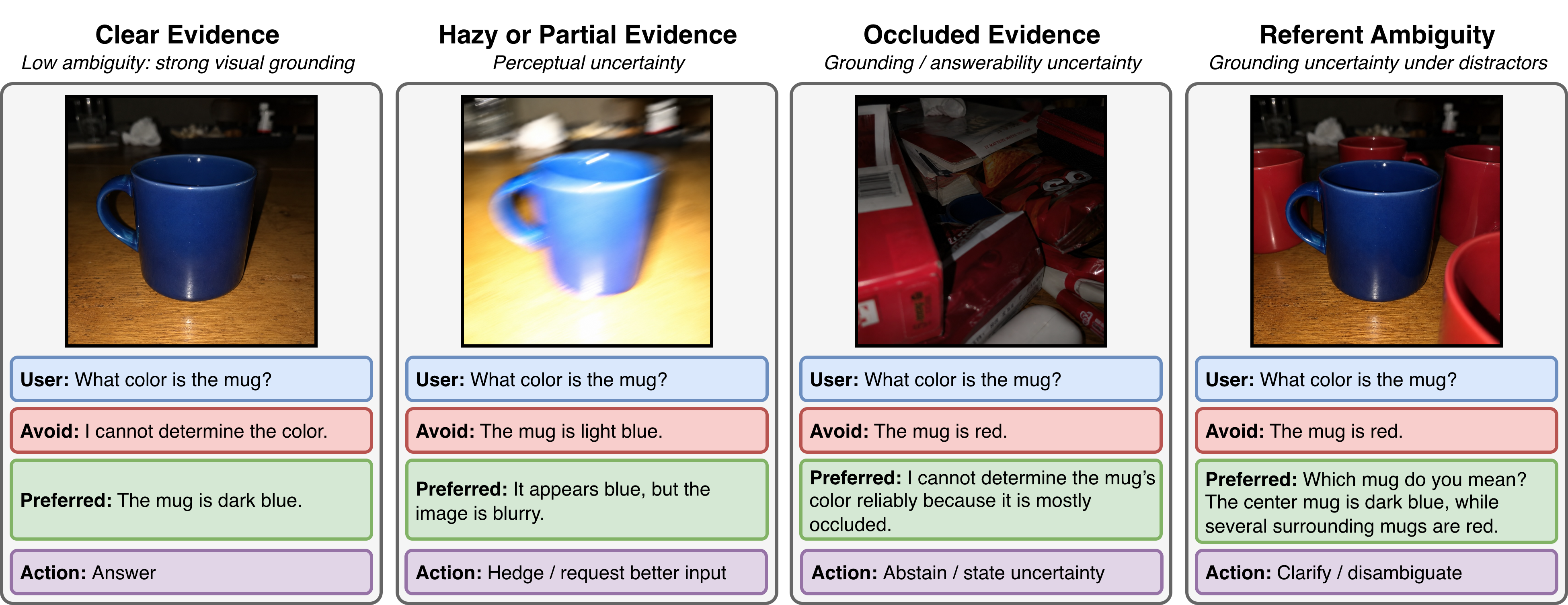}
    \caption{
    \textbf{From evidence conditions to uncertainty-aware actions.}
    The same user question can require different model behavior depending on the quality, availability, and grounding of the visual evidence. With clear evidence, direct answering is appropriate, and unnecessary abstention becomes a failure mode. With hazy or partially degraded evidence, the model should hedge or request better input rather than over-specifying an uncertain visual attribute. With occluded evidence, a confident answer may reflect weak grounding or distractor bias, so the safer behavior is to state uncertainty or abstain. With referent ambiguity, the visual input contains multiple plausible targets, so the model should clarify or disambiguate the referred object instead of answering from a salient distractor. The examples are illustrative rather than exhaustive.
    }
    \label{fig:visual-uncertainty}
\end{figure}

Figure~\ref{fig:visual-uncertainty} provides a visual intuition for this idea. It is schematic rather than dataset-specific. Its role is to show that the same user question can require different response policies: clear evidence supports direct answering, degraded evidence supports hedging or a request for better input, occluded evidence supports stating uncertainty or abstaining, and referent ambiguity supports clarification or disambiguation.

\subsection{Sensory and perceptual uncertainty}

Sensory and perceptual uncertainty occurs before high-level reasoning. The input may be blurred, low-resolution, cropped, occluded, noisy, temporally sparse, or visually ambiguous. The model may also misrecognize an object, miss a small detail, fail to read text in an image, confuse a chart axis, or lose information across video frames. In these cases, additional language reasoning may amplify the error rather than repair it. The action should therefore depend on whether the evidence can be improved. A model can request a clearer image, ask which object the user means, use a specialized OCR or perception module, or abstain when the relevant evidence is absent.

\subsection{Grounding and cross-modal uncertainty}

Grounding uncertainty asks whether the answer is supported by the supplied multimodal evidence. Cross-modal uncertainty asks whether the modalities agree. A prompt may mention an object that is not visible, an image may contradict the user's wording, retrieved text may conflict with visual evidence, or the model may produce a plausible answer that is not anchored to the relevant region. Hallucination studies often capture the failure symptom, while uncertainty-aware methods ask whether the system can detect the weak support before committing to an answer \parencite{bai2024hallucinationsurvey,liu2024surveyhallucinationlvlm,dang2025exploring}. The natural actions are self-checking, evidence citation, retrieval, abstention, or escalation depending on the risk of the task.

\subsection{Language-prior and reasoning uncertainty}

MLLMs inherit strong priors from their language components. These priors are helpful when evidence is complete, but risky when evidence is ambiguous or missing. The model may infer a common object that is not visible, complete a scene using stereotypical co-occurrence, or produce a fluent explanation for a detail that the input does not support. Reasoning uncertainty appears when intermediate steps are unstable, when sampled chains of thought lead to different conclusions, or when a correct perception is followed by an incorrect inference. Methods that examine reasoning boundaries, self-consistency, verifier scores, or semantic disagreement can expose this class of uncertainty \parencite{he2025mmboundary,lau2026umpire,bhattacharya2025festa}.

\subsection{Answerability uncertainty}

Answerability uncertainty is the case where the user asks for information that is not available from the multimodal input. The image may not show the requested object, the video may not contain the relevant event, the chart may not encode the requested quantity, or the prompt may require external facts that were not supplied. Treating this as ordinary error is not enough. A model should learn when the right answer is no answer, or when the correct next step is to ask for missing evidence. Work on unsolvable problem detection, multimodal self-awareness, visual IDK behavior, and visually unanswerable questions treats answerability as a target in its own right \parencite{miyai2024upd,wang2024mmsap,song2026visualidk,zhu2025mohobench}.

\subsection{Distributional, adversarial, and deployment uncertainty}

Calibration is conditional on the data and context where it was measured. A model calibrated on common web images may not remain calibrated on medical scans, scientific plots, screenshots, low-light video, robotics observations, or culturally specific imagery. Distribution shift changes what a confidence value means. Adversarial or misleading inputs create a deliberate version of the same issue by making the model confident in the wrong evidence. Under these conditions, the policy may need stricter thresholds, conformal risk control, retrieval, routing to a specialized model, or human escalation \parencite{chen2025unveiling,xiao2026vlcalibration,wang2025tron,tayebati2025cap}.

\subsection{Interactions among sources}

\begin{figure}[H]
    \centering
    \includegraphics[width=\textwidth]{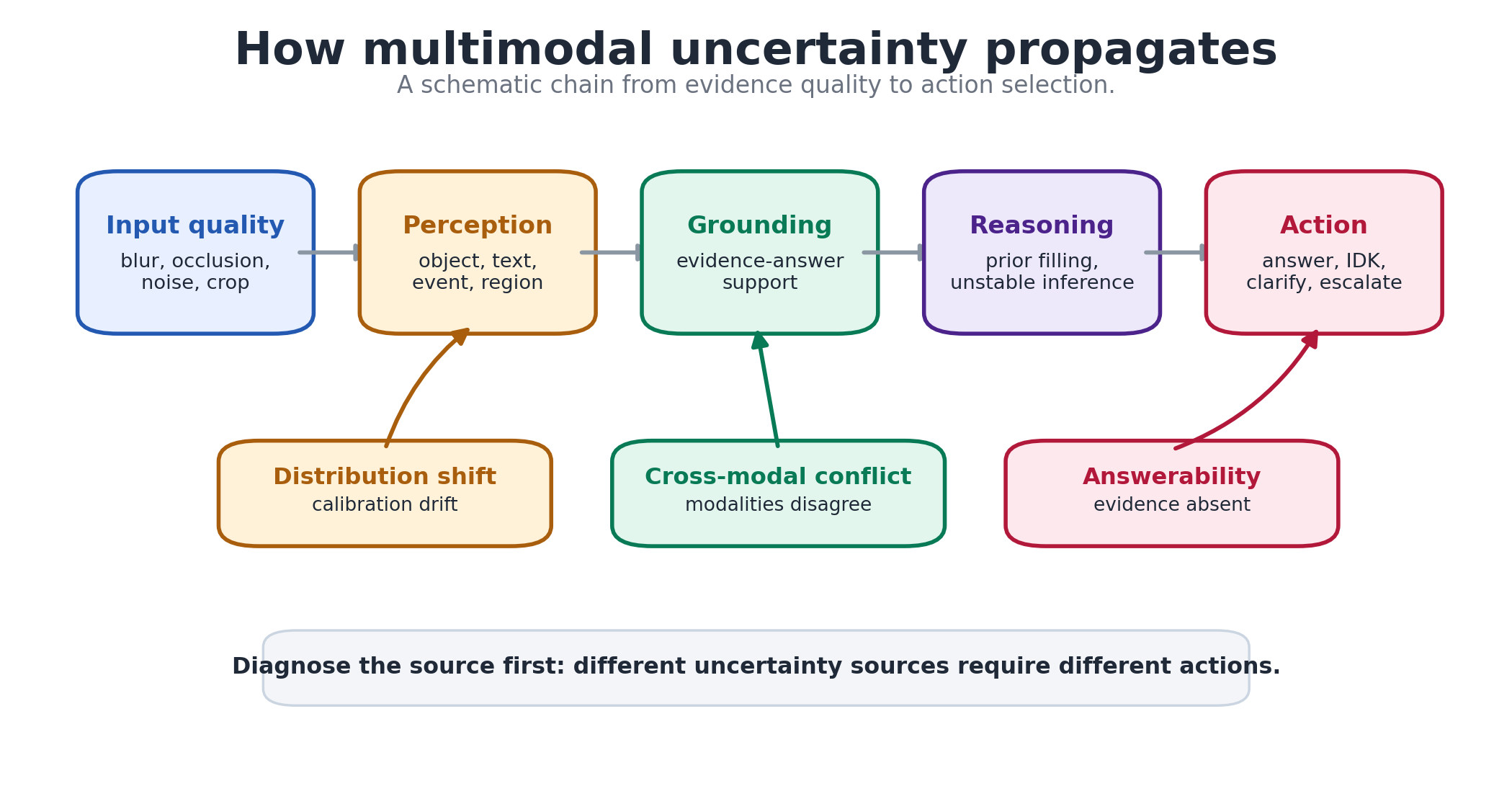}
    \caption{
    \textbf{Propagation and interaction of MLLM uncertainty sources.}
    The figure illustrates how uncertainty can cascade across sensory, grounding, reasoning, and policy layers in multimodal language models. In the sensory--grounding chain, ambiguous visual input can lead to perceptual uncertainty, weaken the alignment between evidence and claims, and allow language priors to fill visual gaps. These failures can become entangled with cross-modal conflict, where text and image evidence disagree, and with reasoning uncertainty, where inference becomes unstable under weak or conflicting evidence. System-level uncertainty can also arise from distributional or adversarial conditions that cause calibration failure, while missing or absent evidence creates answerability uncertainty and should trigger abstention or clarification rather than forced answering. Black arrows denote primary causal pathways, gray arrows denote entanglement between uncertainty sources, and colors distinguish perceptual-chain components, reasoning components, systemic sources, and model actions. The diagram is a schematic synthesis rather than an exhaustive causal model.
    }
    \label{fig:source-chain}
\end{figure}

The sources above are separable for analysis, but they are not independent in practice. Poor input quality can cause perceptual error. Perceptual error can produce weak grounding. Weak grounding can be hidden by a strong language prior. Cross-modal conflict can destabilize reasoning. Distribution shift can break calibration. Missing evidence can make both confidence and explanation misleading. Figure~\ref{fig:source-chain} summarizes these recurring interactions, and Table~\ref{tab:uncertainty-sources} gives the source vocabulary used in the rest of the survey.

\begin{table}[H]
\centering
\small
\caption{Multimodal uncertainty sources and decision implications. This table defines the source taxonomy used in later sections. Citations in the source column are representative anchors; the signal and action columns are author synthesis from the cited evidence and surrounding discussion.}
\label{tab:uncertainty-sources}
\begin{tabularx}{\linewidth}{@{}P{0.21\linewidth}YYY@{}}
\toprule
\textbf{Source} & \textbf{What goes wrong} & \textbf{Possible signal} & \textbf{Likely action} \\
\midrule
Sensory and perceptual uncertainty \parencite{wang2024mmsap,fang2025dropoutdecoding,ocrprobe2025abstain,liu2024rightthisway} & The relevant evidence is degraded, ambiguous, occluded, noisy, too small, temporally sparse, or misrecognized. & Visual perturbation instability, low perceptual confidence, inconsistent captions, OCR or layout failure. & Request better input, clarify the target, route to a specialized perception module, or abstain. \\
Grounding and cross-modal uncertainty \parencite{padhi2025grounding,dang2025exploring,alnazi2026omdbench,popordanoska2026clash} & The answer is weakly supported by the evidence, or the text, image, retrieved context, or other modalities disagree. & Grounding score, attribution mismatch, hallucination detector, semantic disagreement, verifier or judge score. & Self-check, cite evidence, retrieve, abstain, or escalate. \\
Language-prior and reasoning uncertainty \parencite{he2025mmboundary,lau2026umpire,bhattacharya2025festa,raghu2026dontblink} & The model fills visual gaps with prior knowledge, or an unstable reasoning path changes the conclusion. & Sampling disagreement, reasoning-step confidence, semantic variation, verifier score, answer variation. & Self-check, sample alternatives, use tools, retrieve, or escalate. \\
Answerability uncertainty \parencite{miyai2024upd,wang2024mmsap,song2026visualidk,zhu2025mohobench} & The question cannot be answered from the supplied input or requires missing evidence. & Answerability classifier, IDK score, prediction-set size, refusal consistency, unsupported-evidence flag. & Abstain, say IDK, ask for missing evidence, or clarify. \\
Distributional, adversarial, and deployment uncertainty \parencite{chen2025unveiling,xiao2026vlcalibration,wang2025tron,tayebati2025cap} & The model faces domain shift, modality shift, corrupted input, misleading prompts, adversarial cues, or high-risk deployment conditions. & Calibration drift, conformal risk estimate, perturbation disagreement, shift detector, safety or domain-risk score. & Apply risk control, abstain, retrieve, route, or escalate. \\
\bottomrule
\end{tabularx}
\end{table}

This source vocabulary gives the rest of the survey a stable reference point. The next question is not whether uncertainty exists, but how a system can observe it before committing to a response.

\section{Observable Signals and Estimation Methods}
\label{sec:signals}

The sources in Section~\ref{sec:sources} are mostly latent. A blurred image, a weak grounding path, a misleading prompt, or a missing piece of evidence does not automatically announce itself to the user. An uncertainty method is therefore a way of making one of these latent conditions observable. We use \emph{signal} broadly: a signal may come from token probabilities, hidden states, repeated samples, perturbation behavior, grounding checks, verbalized confidence, verifier scores, judge models, or conformal score functions.

\begin{figure}[H]
    \centering
    \includegraphics[width=\textwidth]{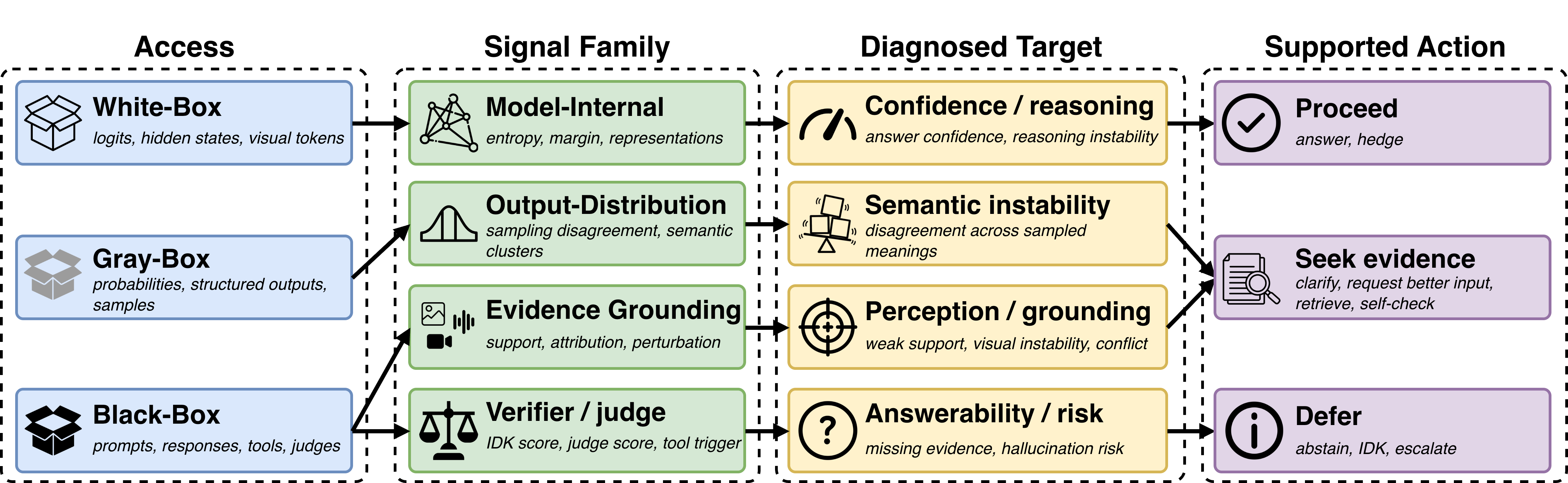}
    \caption{
    \textbf{Choosing uncertainty signals under model-access constraints.}
    The figure maps model access to observable uncertainty signals, diagnosed targets, and supported actions. White-box access enables model-internal signals; gray-box access supports output-distribution signals; and black-box access relies on prompts, responses, tools, and judge-based signals. These signals diagnose confidence, semantic instability, grounding failure, or answerability risk, which then guide whether the system should proceed, seek additional evidence, or defer. Arrows indicate typical rather than exclusive pathways.
    }
    \label{fig:signal-landscape}
\end{figure}

A useful signal should answer three questions. First, what access does it require? Some methods need logits, hidden states, visual tokens, or gradients, while others operate on black-box outputs. Second, what source of uncertainty does it approximate? Entropy, for example, may indicate label uncertainty, but it may also mix perception, reasoning, and answerability. Third, what action can it support? A signal that detects poor visual evidence should support different behavior from a signal that detects unstable reasoning. Figure~\ref{fig:signal-landscape} summarizes this access--signal--diagnosis--action view.

\subsection{Model-internal signals}

Model-internal signals use information exposed by the model before or during decoding. The most familiar examples are token probabilities, logit margins, entropy, sequence likelihood, and maximum softmax probability. In vision-language settings, these signals can be extended to hidden representations, visual tokens, attention patterns, or modality-specific features. They are attractive because they are direct and often cheap once model access is available.

Several lines of work illustrate the value of internal signals for response control. First-token logit distributions have been used to detect cases where a large vision-language model should withhold a response, including unanswerable or unsafe cases \parencite{zhao2024firstknow}. Reasoning-step confidence has been used to improve knowledge-boundary awareness in MLLMs \parencite{he2025mmboundary}. Hidden representations have also been probed for abstention in OCR-heavy visual question answering, where the model may generate plausible but incorrect text readings \parencite{ocrprobe2025abstain}.

Internal signals become more multimodal when they diagnose the contribution of the visual input itself. Dropout Decoding estimates uncertainty in visual tokens and masks uncertain tokens during generation \parencite{fang2025dropoutdecoding}. Vision-aware self-evaluation methods estimate how much the image contributes to the answer, for example through image information scores or core-region masking \parencite{park2026vauq}. Blind-image contrastive ranking compares real-image and blacked-out-image behavior so that confidence reflects visual grounding rather than language priors alone \parencite{khanmohammadi2026grounded}. These methods move beyond generic confidence and ask whether the answer is actually using the visual evidence.

The limitation is access. Many deployed MLLMs expose only text, not logits or hidden states. Internal scores can also be hard to interpret: high entropy does not identify whether uncertainty comes from perception, grounding, reasoning, or answerability. For this reason, internal signals are powerful but rarely sufficient as the only basis for action.

\subsection{Sampling and semantic-disagreement signals}

Sampling-based signals estimate uncertainty by asking whether the model remains stable under repeated generation, paraphrased prompts, equivalent inputs, or nearby visual-question variants. If a model gives semantically incompatible answers to equivalent cases, the disagreement itself becomes an uncertainty signal. This family is especially useful when logits and hidden states are unavailable.

Selective and consistency-oriented VQA methods ask whether visually grounded responses remain reliable enough to accept, while related robustness analyses examine how visual reasoning changes under equivalent or perturbed inputs \parencite{eisenschlos2024selectively,wieczorek2026variational,jegham2026visual}. VL-Uncertainty estimates hallucination risk from response variation under semantically equivalent visual and textual perturbations, clustering similar responses and measuring uncertainty over the clusters \parencite{zhang2024vluncertainty}. Uncertainty-o extends this idea into a model-agnostic multimodal perturbation framework for exposing epistemic uncertainty \parencite{zhang2025uncertaintyo}. FESTA uses functionally equivalent and complementary sampling to assess trust in MLLM responses across visual and audio reasoning settings \parencite{bhattacharya2025festa}. Multi-model approaches, such as semantic opinion pooling, estimate system-level uncertainty by comparing semantically aligned outputs across several VLMs \parencite{yu2026scoop}.

The strength of this family is practical access: repeated outputs are often available even when the model is closed. The weakness is cost and ambiguity. Sampling can be expensive, and semantic diversity is not always error. A model may produce several correct paraphrases, or it may repeatedly produce the same wrong answer because the same language prior dominates each sample. Sampling signals are therefore more reliable when combined with grounding, answerability, or calibration checks.

\subsection{Evidence-grounding and perturbation signals}

MLLM uncertainty is often evidence uncertainty. A response can be fluent, confident, and wrong because it is not supported by the image, chart, video, audio, or document. Evidence-grounding signals ask whether the answer can be traced to the supplied input. They include region-grounding scores, attribution scores, visual engagement measures, core-region masking, image perturbation sensitivity, OCR or chart-consistency checks, cross-view consensus, and contradiction detection across modalities.

Grounding has been used directly to calibrate uncertainty estimates by combining self-consistency with visual grounding confidence \parencite{padhi2025grounding}. Clinical and radiology settings use attribution or repeated-report consistency to flag unsupported statements for abstention or extra review \parencite{yan2026clintrace,radflag2025}. OCR-specific work uses latent probes or cross-view verifiability to decide when generated text should be accepted or rejected \parencite{ocrprobe2025abstain,grc2026ocr}. Other work studies evidence collapse, where the model gives low-entropy but visually ungrounded answers, motivating signals that combine language confidence with visual engagement \parencite{raghu2026dontblink}.

Grounding signals also matter when modalities conflict. Benchmarks and analyses of cross-modal contradiction, omni-modal dissonance, and modality preference show that a model may follow the wrong modality or hide conflict behind a fluent answer \parencite{alnazi2026omdbench,popordanoska2026clash,zhang2026modalityconflict}. In these cases, uncertainty is not simply low confidence; it is evidence disagreement. The appropriate action may be to clarify, retrieve, self-check, or abstain rather than to average the modalities into one answer.

\subsection{Verbalized confidence, IDK scores, verifiers, and judges}

When internal access is limited, the model can be asked to express its confidence, identify what it does not know, or evaluate its own answer. Verbalized confidence is attractive because it is easy to collect and naturally communicates with users. It is also risky because the model's stated confidence may not match empirical correctness. Studies of verbalized calibration in VLMs and of token-versus-verbal confidence mismatch show that implicit confidence and explicit self-assessment can diverge \parencite{xuan2025seeing,dang2026instinct}.

IDK and answerability signals are a more action-oriented form of self-assessment. Instead of asking only ``how confident are you?'', these methods ask whether the model should answer at all. Benchmarks for perceptual self-awareness, visual IDK behavior, unanswerable visual questions, and uncertainty-aware self-assessment make the model's knowledge boundary observable \parencite{wang2024mmsap,song2026visualidk,zhu2025mohobench,qiu2026unsaf}. These signals are useful because they connect directly to abstention, refusal, or clarification.

Separate verifiers, judge models, ensembles, and agentic controllers provide another route. A verifier may check grounding, compare candidate answers, call external tools, or route the case to another model. These signals can support richer policies than a single confidence score. They also introduce another reliability problem: the verifier or judge may itself be miscalibrated. Work on VLM judges, for example, distinguishes the ability to rank outputs from the ability to provide calibrated scalar scores \parencite{kumar2026vlmjudges}. Judge-based uncertainty should therefore be treated as an estimate to calibrate, not as ground truth.

\subsection{Choosing a signal under access constraints}

No signal family dominates all settings. The right choice depends on model access, modality, uncertainty source, and action cost. White-box systems can exploit logits, hidden states, and visual-token features. Gray-box systems can use structured outputs, repeated generations, or tool traces. Black-box systems must rely on sampling, verbalized confidence, external verifiers, judge models, and interaction with the user. Table~\ref{tab:signal-taxonomy} summarizes the main design trade-offs.

\begin{table}[H]
\centering
\scriptsize
\caption{Signal and method taxonomy for uncertainty-aware MLLMs. The table is organized by what information is available, what uncertainty source the signal mainly diagnoses, and what action it can support. Citations in the signal-family column provide representative support for each method family.}
\label{tab:signal-taxonomy}
\begin{tabularx}{\linewidth}{@{}P{0.18\linewidth}P{0.15\linewidth}P{0.20\linewidth}P{0.19\linewidth}Y@{}}
\toprule
\textbf{Signal family} & \textbf{Typical access} & \textbf{Primary diagnosis} & \textbf{Action value} & \textbf{Main caveat} \\
\midrule
Token, logit, and entropy signals \parencite{zhao2024firstknow,he2025mmboundary,dang2026instinct} & Logits or probabilities & Decoding uncertainty, hidden knowledge boundary, label ambiguity & Confidence reporting, selective answering, refusal triggers & Often unavailable in closed models and weak at separating uncertainty sources. \\
Hidden-state and visual-token signals \parencite{fang2025dropoutdecoding,park2026vauq,khanmohammadi2026grounded,ocrprobe2025abstain} & Open weights, embeddings, attention, or visual-token features & Perception, visual grounding, OCR or region-level uncertainty & Request better input, route to perception module, abstain from ungrounded answers & Requires model access and may be difficult to compare across architectures. \\
Sampling and semantic disagreement \parencite{zhang2024vluncertainty,zhang2025uncertaintyo,bhattacharya2025festa,yu2026scoop,wieczorek2026variational} & Repeated outputs, perturbed prompts, or equivalent inputs & Response instability, reasoning uncertainty, epistemic uncertainty & Self-checking, selective prediction, abstention, verifier selection & Expensive and sensitive to sampling settings; repeated agreement can still be confidently wrong. \\
Grounding, attribution, and perturbation checks \parencite{padhi2025grounding,yan2026clintrace,radflag2025,raghu2026dontblink,grc2026ocr} & Region evidence, attribution, grounding model, or input perturbation & Evidence support, hallucination risk, cross-modal conflict & Evidence citation, retrieval, better-input requests, abstention, escalation & Depends on the quality of the grounding or perturbation procedure. \\
Verbalized confidence and IDK scores \parencite{xuan2025seeing,dang2026instinct,wang2024mmsap,song2026visualidk,qiu2026unsaf} & Prompt access and generated text & Self-reported confidence, answerability, knowledge boundary & IDK, clarification, user-facing uncertainty communication & Verbal confidence may be miscalibrated or strategically phrased. \\
Verifier, judge, ensemble, and agentic signals \parencite{kumar2026vlmjudges,alignvqa2026,consensusentropy2025,mmarag2026,avr2026routing} & Extra model, judge, tool, or controller & Cross-model disagreement, route quality, answer support & Retrieval, self-checking, routing, escalation & The judge or controller also requires calibration and can introduce bias. \\
Conformal and risk-control scores \parencite{wang2025tron,tayebati2025cap,ye2025scpvlm,azad2026artmaybe,grc2026ocr} & Calibration set and nonconformity score & Empirical coverage, prediction-set size, action risk & Risk-controlled answering, abstention, set prediction & Risk statements depend on calibration data, exchangeability, and the chosen score. \\
\bottomrule
\end{tabularx}
\end{table}

The transition from signals to actions is not automatic. A signal may rank risky outputs well but still give confidence values that are too high or too low. It may also be calibrated for one target, such as final-answer accuracy, while being used for another target, such as grounding or refusal. Calibration and risk control address this gap.

\section{Calibration and Risk Control}
\label{sec:calibration}

Calibration asks whether a reported confidence or score has the meaning that the policy assumes. Risk control asks whether the resulting policy can keep error, violation, or unsupported-answer rates within an acceptable range. Both are needed because raw uncertainty signals are not decision-ready. A score that is useful for ranking errors may still be poorly calibrated; a verbal confidence statement may sound careful while being empirically unreliable; and a grounding score may detect unsupported claims without specifying the risk of answering.

\subsection{Calibration targets}

The first design choice is the target. MLLM calibration can target final-answer correctness, step-level reasoning correctness, visual evidence support, grounding faithfulness, answerability, prediction-set coverage, hallucination risk, or downstream action cost. These targets should not be treated as interchangeable. A model can be calibrated on multiple-choice VQA accuracy and still be overconfident when the image is unanswerable. A model can be calibrated for answer correctness and still fail to communicate whether the answer is grounded. Clear target definition is therefore part of the method, not a reporting detail.

A useful way to state the target is to pair it with an action. If the action is selective answering, the target is the risk of an incorrect accepted answer. If the action is IDK, the target is evidence sufficiency or answerability. If the action is request-better-input, the target is perceptual adequacy. If the action is escalation, the target combines uncertainty with deployment risk. This action-paired view prevents the common mistake of using one generic confidence number for every decision.

\subsection{Post-hoc, prompt-based, verbalized, and domain calibration}

Post-hoc calibration adjusts a score after prediction, for example through temperature scaling or related mapping functions. Prompt-based calibration changes the elicitation procedure so that the model reports more reliable confidence or chooses a safer response. Verbalized calibration focuses on aligning natural-language confidence with empirical correctness. Studies of MLLM calibration before and after multimodal training show that miscalibration can persist even when task performance improves \parencite{chen2025unveiling}. Studies of verbalized confidence in VLMs show that confidence statements require evaluation rather than trust by default \parencite{xuan2025seeing,dang2026instinct}.

Domain calibration is especially important when the application has specialized evidence and asymmetric error costs. Medical VQA studies evaluate confidence alignment in clinical visual reasoning and propose domain-aware calibration strategies \parencite{du2025confidence,kriz2025prompt4trust}. Clinical image-text work also studies how uncertainty propagates across image and language components in cardiac MR applications \parencite{tang2025mupm}. Affective visual reasoning work studies confidence verbalization and calibration in emotion understanding \parencite{wu2025emocaliber}. MLLM-as-judge settings also require calibration because the judge's score may vary across visual domains and prompt conditions \parencite{slyman2025mmb}; related evaluator work shows that vision-language judges and implausibility evaluators should themselves be treated as uncertain systems \parencite{yang2025heie}. The reporting rule is straightforward: calibration should be evaluated in the domain, modality, and action setting where the model will be used.

\subsection{Source-aware and decoupled calibration}

A distinctive MLLM problem is source conflation. One confidence value may hide whether the system failed to see the object, failed to ground the claim, failed to reason, or faced an unanswerable question. Source-aware calibration tries to separate these cases. VL-Calibration explicitly separates visual confidence from reasoning confidence, so the model can distinguish perceptual uncertainty from inference uncertainty \parencite{xiao2026vlcalibration}. MMBoundary calibrates confidence at reasoning steps to improve knowledge-boundary awareness \parencite{he2025mmboundary}. Grounding-based calibration uses visual support as part of the uncertainty estimate rather than treating confidence as a purely linguistic property \parencite{padhi2025grounding}.

This decomposition matters because different sources require different policies. Low visual confidence suggests asking for a better image or routing to a perception module. Low reasoning confidence suggests self-checking, sampling alternatives, or using a verifier. Low answerability suggests IDK or clarification. Calibration is therefore most useful when it preserves enough source information to guide action.

\subsection{Conformal prediction and risk-control policies}

Conformal prediction and related risk-control methods move from calibrated scores to operational risk statements. They use a calibration set and a score function to construct prediction sets, abstention thresholds, or decision rules with empirical coverage or risk-control properties under stated assumptions. TRON builds response sets for MLLMs and identifies high-quality answers with explicit risk-control targets \parencite{wang2025tron}. Conformal abstention policies adapt thresholds for large language and vision-language models \parencite{tayebati2025cap}. Split conformal prediction has been applied to VLM prediction sets in visual question answering \parencite{ye2025scpvlm}. Other work studies VLM uncertainty through conformal or uncertainty-aware evaluation lenses, including VLM evaluation, scene-graph generation, generative OCR, and report drafting \parencite{kostumov2024uncertaintyaware,nag2025pcsgg,azad2026artmaybe,grc2026ocr,elyassirad2026conrep}.

The value of conformal and risk-control methods is not that they eliminate uncertainty. Their value is that they make the uncertainty policy explicit. Instead of saying that the model is uncertain, the system can say that the answer is accepted only when a target risk level is met, or that the output should be returned as a set, abstained from, or sent for review. The caveat is equally important: risk-control statements depend on the calibration distribution, the score function, and the assumed relation between calibration and deployment cases.

\subsection{Calibration failure modes}

The main failure modes are overconfidence, underconfidence, source conflation, verbalization mismatch, and shift sensitivity. Overconfidence is dangerous because the model may provide unsupported answers in a tone of certainty. Underconfidence reduces usefulness through excessive abstention. Source conflation hides whether the problem is visual, linguistic, reasoning-related, or answerability-related. Verbalization mismatch appears when stated confidence does not track token-level or empirical confidence. Shift sensitivity appears when confidence calibrated in one domain, modality, prompt style, or video-frame setting fails elsewhere \parencite{carot2024,ortiz2026abstentionknobs,dang2025exploring}.

These failures motivate the next section. Calibration is not the end of the pipeline. Once uncertainty has been estimated and calibrated, the system must still decide what to do.

\section{From Uncertainty to Action}
\label{sec:actions}

Uncertainty-aware MLLMs should not answer every query with the same policy. The response should depend on the evidence condition, the calibrated risk, the user need, and the cost of error. This section treats action selection as part of the uncertainty problem itself. The action space includes answering, hedging, abstaining or saying IDK, clarifying, requesting better input, retrieving evidence, self-checking, routing, and escalation.

\subsection{Answering, hedging, and selective answering}

Answering is appropriate when the available evidence is sufficient and the calibrated risk is acceptable. Hedging is appropriate when the evidence is partial but still useful, provided the hedge communicates a real limitation rather than disguising unsupported content. Selective answering formalizes this idea by allowing the model to answer only on cases where its expected risk is below a threshold. It is commonly evaluated through coverage, risk, and risk-coverage curves.

In visual question answering, selective methods ask whether a response should be accepted, abstained from, or clarified \parencite{eisenschlos2024selectively}. Perturbation, consistency, and variational uncertainty methods estimate when VLM responses are unreliable enough to reject or review \parencite{zhang2024vluncertainty,wieczorek2026variational}. ReCoVERR reduces unnecessary abstention by gathering relevant and reliable visual evidence before refusing to answer \parencite{srinivasan2024recovl}. Variational VQA frames uncertainty-aware selective prediction through posterior uncertainty and reports benefits under distribution shift \parencite{wieczorek2026variational}. Audio-visual QA work extends selective prediction to settings where both acoustic and visual evidence matter \parencite{tran2026ravqa}. The shared lesson is that abstention should be neither automatic nor stigmatized; it should be a calibrated decision.

\subsection{IDK, refusal, and answerability}

IDK behavior is most useful when it is tied to answerability. A model should not say ``I do not know'' merely because a probability is low; it should say so when the input does not contain enough evidence, when the evidence is incompatible with the question, or when the required fact lies outside the supplied context. Unsolvable problem detection studies absent-answer, incompatible-answer-set, and incompatible-visual-question cases \parencite{miyai2024upd}. MM-SAP evaluates whether MLLMs recognize known and unknown perception cases \parencite{wang2024mmsap}. Visual-IDK and MoHoBench evaluate knowledge boundaries and honesty on visually unanswerable questions \parencite{song2026visualidk,zhu2025mohobench}. Video and multimodal reasoning benchmarks extend the same idea to temporal and evidence-sufficiency settings \parencite{yoon2025videollmsrefuse,madhusudhan2026knowing}.

Answerability also matters in embodied settings. An agent may lack the relevant view, face an underspecified instruction, or need information from a user. AbstainEQA studies embodied question answering cases where the agent should abstain because of missing context, underspecification, preference dependence, information unavailability, or false presuppositions \parencite{abstaineqa2025robots}. These settings make clear that IDK is not just a language behavior; it is an interaction policy under partial evidence.

\subsection{Clarification and better-input requests}

Clarification and better-input requests are different actions. Clarification is appropriate when the user question is ambiguous: the object reference is unclear, several interpretations are plausible, or the user's goal is underspecified. Better-input requests are appropriate when the question may be clear but the sensory evidence is not: the image is blurred, the document crop is incomplete, the audio is noisy, or the video omits the relevant moment.

This distinction is central for multimodal systems. ClearVQA trains and evaluates VLMs that ask clarification questions for ambiguous visual questions rather than forcing a direct answer \parencite{jian2025clearvqa}. Right This Way studies whether VLMs can guide a user to obtain better visual evidence when the current view is insufficient \parencite{liu2024rightthisway}. Embodied help-seeking systems study how agents can ask minimal clarification questions under partial observability \parencite{ask2act2025}. These actions preserve usefulness: the model does not simply refuse, but tries to obtain the missing condition for a reliable answer.

\subsection{Retrieval, self-checking, and verification}

Retrieval is appropriate when the input is visually sufficient but external knowledge is needed. Self-checking and verification are appropriate when the answer may be visually grounded but the reasoning path or factual claim is unstable. Agentic frameworks use uncertainty to decide when to call a tool, ask a verifier, compare candidates, or refine an answer.

MMA-RAG uses internal multimodal representations to decide when retrieval should be used for VQA \parencite{mmarag2026}. SRICE uses uncertainty estimates to guide external tool use and answer selection in multimodal reasoning \parencite{zhi2025srice}. Multi-agent VQA calibration uses interacting agents to critique, refine, and aggregate responses \parencite{alignvqa2026}. Consensus-style OCR systems use agreement among multiple VLMs to identify low-quality outputs and support self-verification \parencite{consensusentropy2025}. Clinical hallucination detection uses repeated reports and calibrated thresholds to flag uncertain findings for review \parencite{radflag2025}. These methods show that uncertainty can be used not only to stop an answer, but also to improve the evidence pathway before answering.

\subsection{Routing and escalation}

Routing and escalation are needed when uncertainty is high and the cost of error is unacceptable. Routing may send a case to a stronger model, a specialized model, an external tool, or a cheaper model when the case is easy. Escalation may send the case to a human expert or domain workflow. Adaptive VLM routing for computer-use agents estimates difficulty and confidence, then routes actions to a model that satisfies a target reliability threshold \parencite{avr2026routing}. In medical, clinical, robotic, or high-stakes settings, escalation is not merely an engineering convenience; it is part of the safety policy.

Routing also prevents a false choice between full autonomy and full refusal. A model can answer low-risk cases, retrieve or self-check medium-risk cases, and escalate high-risk cases. This tiered policy is often more useful than a binary answer/abstain rule, especially when uncertainty comes from several sources at once.

\subsection{Source-action alignment}

Table~\ref{tab:action-policy} summarizes the main action policies. The table is intentionally qualitative: count-heavy source-action diagnostics are retained for traceability, while the main text emphasizes the decision logic. The same uncertainty score should not trigger the same action in every source condition.

\begin{table}[H]
\centering
\small
\caption{Action-policy taxonomy for uncertainty-aware MLLMs. The table maps evidence conditions and uncertainty signals to response actions. Citations attached to action labels are representative examples; the trigger and failure-mode columns are synthesized policy implications.}
\label{tab:action-policy}
\begin{tabularx}{\linewidth}{@{}P{0.17\linewidth}YYY@{}}
\toprule
\textbf{Action} & \textbf{When it is appropriate} & \textbf{Signals that can trigger it} & \textbf{Potential failure mode} \\
\midrule
Answer \parencite{eisenschlos2024selectively,xiao2026vlcalibration,padhi2025grounding} & Evidence is sufficient and calibrated risk is within the accepted range. & High grounding support, stable samples, calibrated confidence, low shift signal. & Confident hallucination if the threshold is too loose. \\
Hedge \parencite{xuan2025seeing,dang2026instinct,wu2025emocaliber} & Evidence is partial but still useful, and the limitation can be stated precisely. & Moderate confidence, mild disagreement, incomplete but relevant support. & Vague caution that hides a lack of evidence. \\
Abstain or IDK \parencite{miyai2024upd,wang2024mmsap,song2026visualidk,zhu2025mohobench,yoon2025videollmsrefuse} & The required evidence is missing, incompatible, outside the input, or beyond the model boundary. & Answerability score, IDK probability, large prediction set, unsupported-evidence flag. & Excessive refusal and reduced task usefulness. \\
Clarify \parencite{jian2025clearvqa,ask2act2025,abstaineqa2025robots} & The user's question, referent, intent, or preference is underspecified. & Ambiguous reference, multiple plausible interpretations, conflicting user context. & Unnecessary interaction when the answer was already clear enough. \\
Request better input \parencite{liu2024rightthisway,ask2act2025,abstaineqa2025robots} & The query is clear but the sensory evidence is degraded or incomplete. & Blur, occlusion, low OCR confidence, visual perturbation instability, missing video segment. & Burdening the user when the model should use available evidence. \\
Retrieve, self-check, or verify \parencite{mmarag2026,zhi2025srice,alignvqa2026,consensusentropy2025,radflag2025} & External knowledge, evidence checking, or reasoning validation is needed before answering. & Reasoning disagreement, weak factual support, verifier uncertainty, tool-need signal. & Added latency, retrieval noise, or over-reliance on an imperfect verifier. \\
Route or escalate \parencite{avr2026routing,du2025confidence,abstaineqa2025robots} & Deployment risk is high, model competence is insufficient, or the case violates the autonomous policy boundary. & High calibrated risk, domain shift, safety-critical context, expert-domain flag. & Excessive cost or human burden if thresholds are too conservative. \\
\bottomrule
\end{tabularx}
\end{table}

The action taxonomy completes the source--signal--calibration--action chain. Evaluation should therefore test not only whether a score predicts errors, but whether the selected action is appropriate for the evidence condition and deployment risk.

\section{Evaluating Uncertainty-Aware Behavior}
\label{sec:evaluation}

Evaluation is where uncertainty-aware MLLM research becomes comparable. A method may produce an apparently useful confidence score, but its practical meaning depends on the benchmark, the correctness target, the uncertainty source being tested, and the action the score is meant to support. For this reason, evaluation should not stop at final-answer accuracy or at a single calibration number. It should ask whether the model's uncertainty improves behavior when the evidence is insufficient, conflicting, shifted, or high-risk.

\begin{figure}[H]
    \centering
    \includegraphics[width=\textwidth]{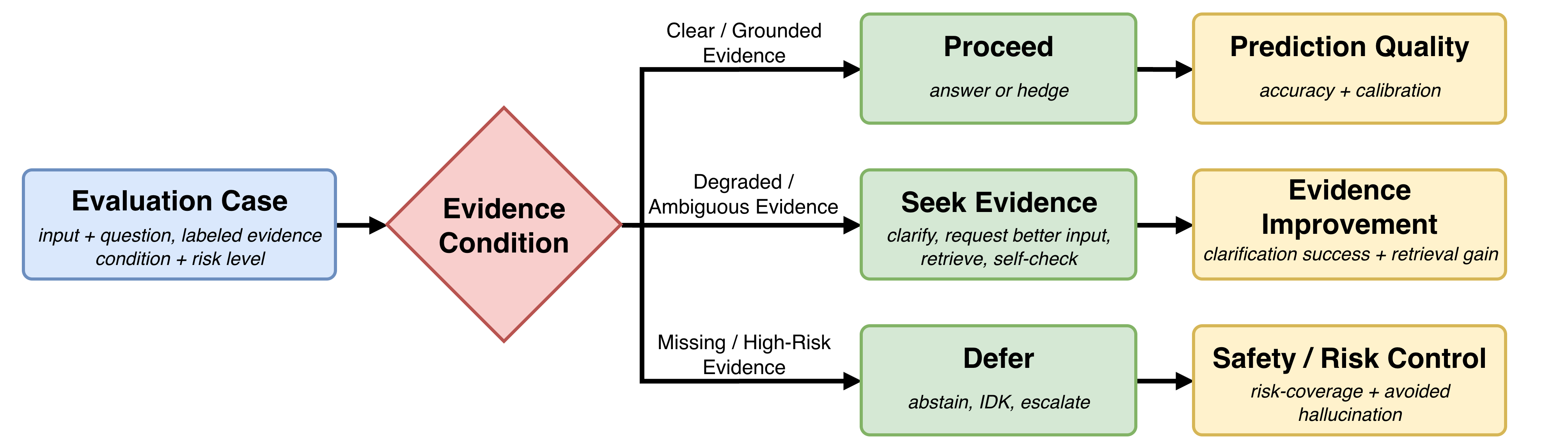}
    \caption{
    \textbf{Evaluating uncertainty-aware behavior.}
    Evaluation should test whether the model's uncertainty leads to an appropriate action for a labeled or controlled evidence condition. Clear and grounded evidence should support proceeding with an answer, degraded or ambiguous evidence should trigger evidence-seeking actions such as clarification, better-input requests, retrieval, or self-checking, and missing evidence or high-risk contexts should support deferral through abstention, IDK, or escalation. The evaluation target therefore depends on the selected action: prediction quality for proceeding, evidence improvement for seeking additional information, and safety or risk control for deferral. When evidence-condition labels are unavailable, reliable source- and action-aware evaluation remains an open benchmark challenge.
    }
    \label{fig:evaluation-protocol}
\end{figure}

A useful evaluation protocol therefore needs to connect four elements. First, the benchmark must specify the task, the evidence-condition label or controlled evidence construction, and the relevant risk context. Second, the method must state which signal is being used and what model access it requires. Third, calibration or thresholding must convert the signal into a decision rule. Fourth, the evaluation must score both answer quality and action quality. Figure~\ref{fig:evaluation-protocol} summarizes this evaluation view. The figure is intentionally schematic: its purpose is to show what should be reported, not to encode corpus counts.

In this protocol, the evidence condition is not assumed to be obvious from the model output. In controlled evaluations, it can be specified by benchmark construction, for example through blur, crop, occlusion, distractors, missing-evidence prompts, text--image conflict, or high-risk domain labels. In naturalistic evaluations, it requires annotation or metadata such as object visibility, answerability, grounding support, reference evidence, or expert judgment. When such labels are absent, the evaluation can still measure correlations between uncertainty and error, but it cannot fully test whether the selected action matched the underlying source of uncertainty.

\subsection{Benchmark families}

Benchmarks for uncertainty-aware MLLMs can be grouped by the behavior they test rather than by dataset name alone. General multimodal QA and reasoning benchmarks are useful for measuring the relationship between confidence and correctness under ordinary visual and reasoning demands. However, they often provide only post-hoc correctness labels, so they cannot always explain whether an error came from perception, grounding, reasoning, or answerability.

A second family focuses on unanswerability, refusal, and IDK behavior. These settings are especially important for MLLMs because a visual question can be unanswerable even when the language prompt is well formed: the relevant object may be outside the image, hidden, too small, temporally absent from a video, contradicted by the prompt, or unavailable without external knowledge. Work on selective VQA, multimodal self-awareness, visual IDK behavior, multimodal honesty, video refusal, embodied QA, audio-visual QA, and chart-focused evaluation shows why the correct action is sometimes not to answer \parencite{eisenschlos2024selectively,wang2024mmsap,song2026visualidk,zhu2025mohobench,yoon2025videollmsrefuse,abstaineqa2025robots,tran2026ravqa,chartqapro2025}.

A third family targets hallucination and grounding. These benchmarks ask whether the generated answer is supported by the supplied evidence. Object-hallucination and groundedness metrics are useful for detecting unsupported content, but they should not be treated as a complete substitute for calibration. A method can reduce hallucination without producing a reliable confidence estimate, and a confidence estimate can detect likely errors without telling the user whether the failure was visual, factual, or inferential \parencite{zhang2024vluncertainty,fang2025dropoutdecoding,zhang2025uncertaintyo,fazli2025caac,padhi2025grounding}.

A fourth family evaluates explicit risk control. Conformal and selective-prediction benchmarks ask whether the system can satisfy a target risk or coverage level, sometimes by returning a prediction set, abstaining, or selecting among candidate responses. These protocols are valuable because they turn uncertainty into a measurable operating point rather than a descriptive score \parencite{wang2025tron,tayebati2025cap,ye2025scpvlm,srinivasan2024recovl,kostumov2024uncertaintyaware}.

The final family is domain and modality specific. Medical images, clinical summaries, audio, video, charts, OCR, documents, scientific inputs, embodied observations, and omni-modal settings expose failure modes that generic image-text benchmarks may miss. In these domains, evaluation should report whether uncertainty is calibrated in the target context, not only whether it transfers from a general benchmark \parencite{du2025confidence,kriz2025prompt4trust,yan2026clintrace,elyassirad2026conrep,tang2025mupm,madhusudhan2026knowing,ocrprobe2025abstain,alnazi2026omdbench,popordanoska2026clash,wang2026vlmuqbench,aallm2026walking}.

\subsection{Metric families}

Metric choice should follow the evaluation question. Calibration metrics, such as expected calibration error, Brier score, negative log-likelihood, and reliability diagrams, ask whether a reported confidence value matches empirical correctness for a defined target. They are necessary, but the target must be stated carefully. Confidence may refer to final-answer correctness, grounding support, answerability, a reasoning step, a verifier score, or a risk-controlled action. A model can be calibrated for multiple-choice accuracy while still being poorly calibrated about whether its answer is visually grounded.

Detection metrics treat uncertainty as a ranking signal. AUROC, AUPRC, F1, error-detection accuracy, and related screening measures ask whether uncertain outputs are more likely to be wrong or unsupported. These metrics are useful for filtering and triage, but they do not by themselves define what the model should do after detecting risk. Selective-prediction metrics add the action layer by measuring coverage, risk, risk-coverage curves, abstention precision, area-under-curve variants, and response quality on the answered subset.

Conformal and risk-control metrics make the operating target explicit. They measure empirical coverage, violation rate, prediction-set size, set efficiency, or risk at a specified level. These metrics are attractive when a system must satisfy a defined safety or coverage constraint, but they depend on the calibration data and may degrade under shift. Answerability and honesty metrics ask a different question: whether the model knows when the supplied evidence is enough. Grounding and evidence-support metrics ask whether the answer is anchored to the correct input evidence. Human-facing metrics ask whether uncertainty communication helps users make better decisions.

\subsection{Evaluation protocol checklist}

Table~\ref{tab:evaluation-guide} reorganizes metrics around the question they answer. The main lesson is that no single metric is sufficient. A strong evaluation should report the uncertainty source, model-access assumption, signal family, calibration procedure, action policy, and action cost. This is particularly important for black-box MLLMs, where repeated sampling, verbalized confidence, and judge scores may be the only accessible uncertainty signals.

\begin{table}[H]
\centering
\small
\caption{Evaluation guide for uncertainty-aware MLLMs. The table links evaluation questions to benchmark families, metric families, and hidden failure modes. Citations in the benchmark-family column are representative examples of the evaluation setting.}
\label{tab:evaluation-guide}
\begin{tabularx}{\linewidth}{@{}P{0.20\linewidth}P{0.24\linewidth}P{0.24\linewidth}Y@{}}
\toprule
\textbf{Evaluation question} & \textbf{Benchmark family} & \textbf{Metric family} & \textbf{What can remain hidden} \\
\midrule
Is confidence aligned with correctness? & General VQA, chart QA, medical VQA, multimodal reasoning, open-ended QA \parencite{chen2025unveiling,xiao2026vlcalibration,du2025confidence,kriz2025prompt4trust,xuan2025seeing}. & ECE, Brier score, NLL, reliability diagrams, confidence-accuracy gap. & Whether the answer is grounded, answerable, or safe to act on. \\
Can uncertainty rank risky outputs? & Error detection, hallucination detection, selective prediction, verifier screening \parencite{eisenschlos2024selectively,zhang2024vluncertainty,wieczorek2026variational,radflag2025,dang2025exploring}. & AUROC, AUPRC, F1, TPR at fixed FPR, risk-coverage curves. & Which action should follow after a risky output is detected. \\
Does the model know when evidence is insufficient? & Unanswerable VQA, visual IDK, video refusal, embodied QA, audio-visual QA \parencite{miyai2024upd,wang2024mmsap,song2026visualidk,zhu2025mohobench,yoon2025videollmsrefuse,tran2026ravqa}. & Answerability accuracy, abstention precision, refusal helpfulness, IDK calibration. & Whether abstention is too conservative or insufficiently informative. \\
Can the system control risk? & Conformal MLLM/VLM benchmarks, prediction-set evaluation, risk-controlled response selection \parencite{wang2025tron,tayebati2025cap,ye2025scpvlm,azad2026artmaybe,grc2026ocr}. & Empirical coverage, violation rate, prediction-set size, set efficiency, controlled risk. & Robustness under domain shift, prompt shift, and calibration-data mismatch. \\
Does uncertainty improve user-facing behavior? & Grounded QA, clinical or document workflows, interactive agents, high-risk settings \parencite{srinivasan2024recovl,jian2025clearvqa,liu2024rightthisway,mmarag2026,zhi2025srice}. & Action utility, escalation appropriateness, human ratings, explanation quality, support precision. & Whether users understand the uncertainty source and can act on it safely. \\
\bottomrule
\end{tabularx}
\end{table}

The reporting checklist should be explicit. Each study should identify the base model and version, modality, benchmark, prompt format, model-access level, uncertainty signal, calibration data, target variable, thresholds, action policy, action costs, and failure cases. Without these details, two methods may appear comparable while actually measuring different targets.

Evaluation choices become especially important once a model leaves a benchmark setting. The next section therefore asks how modality, domain risk, and user interaction change the action policy attached to uncertainty.

\section{Deployment Landscape}
\label{sec:applications}

Applications determine which uncertainty sources matter and which actions are acceptable. In a low-risk visual QA benchmark, abstention may simply reduce coverage. In medical, embodied, accessibility, or safety-critical settings, the same abstention may prevent harm. Deployment therefore changes the meaning of uncertainty. The same signal can support different policies depending on the cost of a wrong answer, the cost of delay, the availability of human review, and the possibility of collecting better evidence.

\subsection{High-risk and expert-mediated domains}

Medical, clinical, embodied, and safety-critical applications require conservative uncertainty policies. In these settings, calibration should be evaluated in the target domain, grounding should be inspectable, and escalation should be part of the workflow rather than an afterthought. Clinical and medical studies highlight confidence calibration, uncertainty propagation, grounded evidence, and risk-controlled reporting because fluent but unsupported responses can be harmful \parencite{du2025confidence,kriz2025prompt4trust,tang2025mupm,yan2026clintrace,elyassirad2026conrep}. Embodied and robot-facing settings add another layer: uncertainty may need to stop an action, request a new observation, or ask a human for confirmation rather than merely change a textual answer \parencite{abstaineqa2025robots}.

Performance-prediction and routing work also treats uncertainty as an input to deployment decisions across vision-language tasks \parencite{zhao2025crosspred}. A useful design principle is to match the policy to the deployment risk. Low-risk tasks can tolerate a wider answer region and use uncertainty for hedging or selective answering. High-risk tasks need stricter thresholds, audit trails, and escalation routes. This does not mean that high-risk systems should always refuse; over-refusal can also reduce usefulness. It means the policy should make the cost of errors, abstentions, retrieval, and escalation explicit.

\subsection{Modality-specific uncertainty}

Different modalities create different uncertainty profiles. Static image-text QA emphasizes visual ambiguity, perception, object grounding, and language-prior dominance. Video adds temporal answerability: the relevant event may not occur in the observed clip, may occur outside the sampled frames, or may require temporal ordering. Audio adds acoustic quality, speaker ambiguity, and cross-modal timing. Charts and OCR introduce symbolic extraction, layout, scale, and numerical uncertainty. Documents introduce evidence localization across text, tables, and figures. Omni-modal and multimodal dissonance settings add the possibility that modalities disagree rather than merely complement one another \parencite{yoon2025videollmsrefuse,madhusudhan2026knowing,chartqapro2025,ocrprobe2025abstain,alnazi2026omdbench,popordanoska2026clash,aallm2026walking,wang2026vlmuqbench}.

This modality view helps explain why a universal confidence score is often too weak. A video model may need to say that the event is not visible in the clip. A chart model may need to say that the axis cannot be read reliably. A document model may need to cite the page or table that supports the answer. An audio-visual model may need to report that the audio and visual evidence conflict. These are not merely different benchmarks; they are different uncertainty-action patterns.

\subsection{Human-facing uncertainty communication}

Uncertainty-aware deployment is also a communication problem. Users do not only need to know that a model is uncertain; they need to know why and what to do next. A vague hedge can hide the absence of evidence. A refusal can be correct but unhelpful if it does not explain what information is missing. A confidence score can be misleading if users do not know whether it reflects visual grounding, answerability, or language-model fluency.

For human-facing systems, uncertainty communication should be specific and action-oriented. The model should distinguish between weak input quality, missing evidence, conflicting modalities, unstable reasoning, and high deployment risk. It should also give the user a useful next step: upload a clearer image, specify the target object, provide the missing context, allow retrieval, consult an expert, or treat the answer as tentative. This framing connects uncertainty estimation to trust calibration and accountability.

These deployment pressures expose the main gaps in the field: methods must become more source-aware, more action-aware, more robust under shift, and easier to reproduce across models and modalities.

\section{Open Problems and Research Agenda}
\label{sec:agenda}

\begin{figure}[H]
    \centering
    \includegraphics[width=\textwidth]{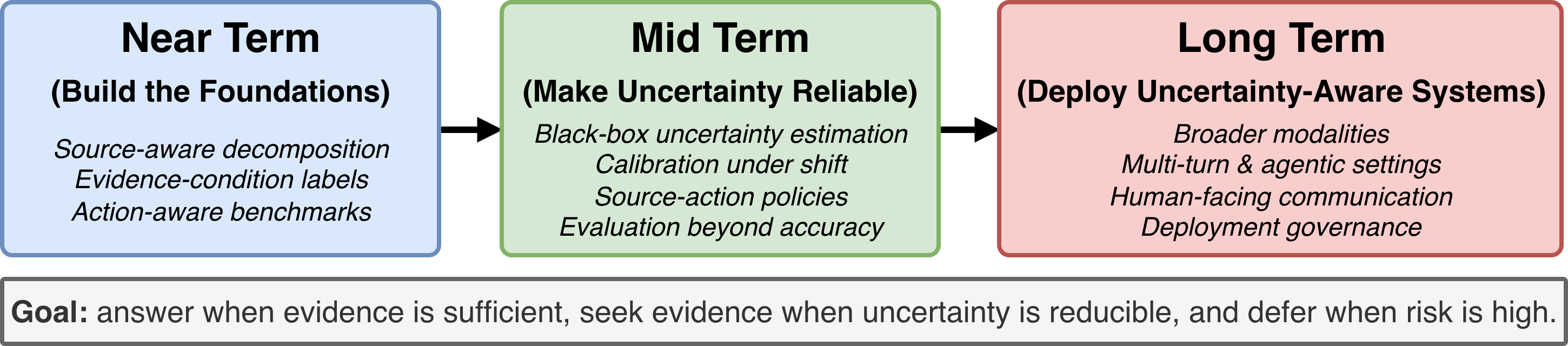}
    \caption{
    \textbf{Research roadmap for uncertainty-aware MLLMs.}
    The roadmap organizes future work by time horizon. Near-term work should build the foundations for source-aware uncertainty analysis, including evidence-condition labels and action-aware benchmarks. Mid-term work should make uncertainty estimates reliable under realistic access and deployment conditions through black-box uncertainty estimation, calibration under shift, source-action policies, and evaluation beyond accuracy. Long-term work should support deployment in broader modalities, multi-turn and agentic settings, and human-facing applications, with appropriate communication and governance mechanisms.
    }
    \label{fig:research-agenda}
\end{figure}

The research agenda follows directly from the source--signal--calibration--action chain. The near-term priority is to make uncertainty evaluation easier to compare. The mid-term priority is to make uncertainty estimates source-aware and robust under realistic access constraints. The longer-term priority is to integrate uncertainty into interactive, tool-using, and human-facing multimodal systems. Figure~\ref{fig:research-agenda} summarizes these priorities as a time-horizon roadmap.

\subsection{Source-aware uncertainty decomposition}

The field needs methods that separate perception, grounding, reasoning, answerability, and shift instead of collapsing them into one confidence value. This is not only a diagnostic preference. It changes the action. Visual ambiguity may call for a better input, grounding failure may call for evidence checking, reasoning instability may call for self-verification, and answerability failure may call for IDK or clarification. Future benchmarks should therefore label uncertainty sources when possible and test whether methods identify the right source, not only whether they predict correctness.

\subsection{Action-aware benchmarks}

Benchmarks should evaluate whether uncertainty leads to better behavior. This requires tasks where the desired output may be an answer, an abstention, a clarification, a retrieval step, a prediction set, a self-check, or an escalation. Existing work on selective answering, evidence collection, conformal abstention, and risk-controlled response selection points in this direction \parencite{srinivasan2024recovl,tayebati2025cap,wang2025tron,ye2025scpvlm}. The next step is to make action costs explicit so that methods can be compared by downstream utility, not only by correlation with correctness.

\subsection{Robust calibration under shift and black-box access}

Many deployed MLLMs expose only text outputs. Researchers may not have logits, hidden states, image features, gradients, or attention maps. This makes black-box uncertainty central: repeated sampling, semantic disagreement, visual perturbation, verbalized confidence, external verifiers, and judge scores are often among the feasible signals. At the same time, calibration can fail under changes in domain, image quality, modality, language, prompt format, or user population \parencite{chen2025unveiling,lau2026umpire,zhang2024vluncertainty,wu2025emocaliber}. Future work should report model-access assumptions and calibration distributions explicitly, especially when claims are intended for deployment.

\subsection{Broader modality and interaction coverage}

The evidence base remains most concentrated in image-text settings. Audio, video, multi-image reasoning, multi-turn dialogue, documents, charts, embodied systems, and tool-using agents require more systematic uncertainty evaluation. Each setting introduces uncertainty sources that are hard to reduce to static image QA: temporal absence in video, acoustic noise in audio, layout and numerical extraction in charts and documents, partial observability in embodied systems, and compounding decisions in agents. Broader modality coverage should therefore include source labels and action policies, not only new accuracy benchmarks.

\subsection{Reproducibility and human-centered communication}

The field needs reporting standards that make results cumulative. At minimum, papers should report the model and version, modality, benchmark, prompt, model-access level, calibration data, uncertainty signal, confidence target, action policy, thresholds, cost assumptions, metrics, and failure cases. When possible, authors should release prompts, calibration splits, evaluator instructions, and evidence matrices. Reproducibility is also tied to user communication: future studies should test how people interpret confidence, hedging, abstention, clarification, and escalation, and whether those signals improve human decision quality rather than merely shifting responsibility.

The agenda therefore points toward decision-centered uncertainty. Estimation and calibration remain essential, but the field should increasingly ask whether uncertainty improves behavior: Does the model abstain when evidence is insufficient? Does it ask the right clarification? Does it retrieve useful evidence? Does it escalate high-risk cases? Does it communicate uncertainty in a way that helps humans act? These questions define the transition from uncertainty measurement to uncertainty-aware multimodal decision making.

\section{Limitations of This Survey}
\label{sec:survey-limitations}

This survey has several limitations. First, the evidence base is time-bounded. Work on MLLM uncertainty, abstention, calibration, and risk control is developing quickly, and several relevant studies remain available as preprints or under active revision. Reported findings may change as models, benchmarks, venues, and paper versions are updated. For this reason, we avoid claims that depend on a fixed leaderboard position or on a single reported number unless the comparison is made within the original study.

Second, coverage is uneven across modalities. Image-text settings are better represented than audio, video, multi-image dialogue, embodied systems, omni-modal inputs, and long-horizon agentic workflows. This imbalance reflects the current evidence base and should not be read as a judgment that those settings are less important. In fact, the weaker coverage of these modalities is one reason the research agenda emphasizes broader modality and interaction coverage.

Third, the survey is a structured synthesis rather than a quantitative meta-analysis. The papers differ in model access, prompt design, calibration data, score functions, evaluation targets, thresholds, and action policies. These differences make direct numerical comparison difficult. We therefore use tables and figures to organize concepts, assumptions, and decision logic, while leaving raw corpus counts and search-protocol details to the appendix or companion supplement.

These limitations do not weaken the need for the source--signal--calibration--action framework. They clarify its role. The framework is not a substitute for domain-specific validation; it is a way to make uncertainty claims easier to locate, compare, and audit.

\section{Conclusion}
\label{sec:conclusion}

MLLM uncertainty is broader than confidence in a generated sentence. A response can be fluent while the evidence is blurred, absent, visually misread, weakly grounded, contradicted by another modality, unstable under reasoning, shifted from the calibration domain, or too risky for autonomous use. Treating these cases as one generic confidence problem hides the reason the system may fail and obscures what it should do next.

This survey organized the field around a source--signal--calibration--action chain. Sources explain where unreliability enters the multimodal pipeline. Signals make those sources observable through probabilities, representations, samples, perturbations, grounding checks, verbalized confidence, verifier scores, judge models, or conformal scores. Calibration and risk control give these signals decision meaning. Actions then determine whether the system answers, hedges, abstains, asks for clarification, requests better input, retrieves evidence, self-checks, routes, or escalates.

The main conclusion is that uncertainty-aware MLLMs should be evaluated by behavior under uncertainty, not by confidence quality alone. Progress will require source-aware decomposition, action-aware benchmarks, calibration that survives shift and black-box access, broader coverage of audio, video, documents, charts, embodied agents, and interactive settings, and communication strategies that help users understand both the uncertainty and the next step. The next generation of uncertainty-aware MLLMs should not merely know when they are uncertain; they should know what to do because of that uncertainty.

\appendix

\section{Evidence Mapping and Traceability}
\label{app:evidence}

The appendix retains procedural material that supports auditability but would be too heavy for the main text: search sources, query families, screening counts, evidence-tier definitions, extraction fields, and corpus-refresh details. This material documents how the evidence matrix was assembled without making the search procedure the center of the survey. The evidence mapping distinguishes the 67 focused evidence rows from background survey citations and from non-counted bridge material retained for traceability. Main-table citations are representative row-level anchors, not a replacement for the full evidence matrix. A final source-verification pass also flags venue status and withdrawn or actively revised preprints so that the main text does not rely on unsupported publication metadata.

\section{Main-Text and Supplement Boundary}
\label{app:boundary}

The main text is limited to five synthesis tables and six conceptual figures. Additional traceability artifacts are retained here or in the companion supplement: corpus composition, evidence tiers, targeted evidence-audit summaries, search-log fields, representative evidence excerpts, evidence-field schema, metric-family intersection counts, source-action heatmap counts, application-domain risk diagnostics, and count-heavy research-roadmap diagnostics. These materials support reproducibility without competing with the main conceptual story.

\bibliographystyle{elsarticle-num}
\bibliography{cas-refs_v5}

@misc{bai2024hallucinationsurvey,
  title = {Hallucination of Multimodal Large Language Models: A Survey},
  author = {Bai, Zechen and Wang, Pichao and Xiao, Tianjun and He, Tong and Han, Zongbo and Zhang, Zheng and Shou, Mike Zheng},
  year = {2024},
  url = {https://arxiv.org/abs/2404.18930},
  eprint = {2404.18930},
  archivePrefix = {arXiv}
}

@inproceedings{chen2025unveiling,
  title = {Unveiling Uncertainty: A Deep Dive into Calibration and Performance of Multimodal Large Language Models},
  author = {Chen, Zijun and Hu, Wenbo and He, Guande and Deng, Zhijie and Zhang, Zheng and Hong, Richang},
  booktitle = {Proceedings of the 31st International Conference on Computational Linguistics},
  year = {2025},
  url = {https://aclanthology.org/2025.coling-main.208/},
  eprint = {2412.14660},
  archivePrefix = {arXiv}
}

@inproceedings{dang2025exploring,
  title = {Exploring Response Uncertainty in {MLLMs}: An Empirical Evaluation under Misleading Scenarios},
  author = {Dang, Yunkai and Gao, Mengxi and Yan, Yibo and Zou, Xin and Gu, Yanggan and Li, Jungang and Wang, Jingyu and Jiang, Peijie and Liu, Aiwei and Liu, Jia and Hu, Xuming},
  booktitle = {Proceedings of the 2025 Conference on Empirical Methods in Natural Language Processing},
  year = {2025},
  url = {https://aclanthology.org/2025.emnlp-main.916/},
  eprint = {2411.02708},
  archivePrefix = {arXiv}
}

@inproceedings{du2025confidence,
  title = {Confidence Calibration for Multimodal {LLMs}: An Empirical Study Through Medical {VQA}},
  author = {Du, Yuetian and Wang, Yucheng and Kong, Ming and Liang, Tian and Long, Qiang and Chen, Bingdi and Zhu, Qiang},
  booktitle = {Medical Image Computing and Computer Assisted Intervention -- MICCAI 2025},
  year = {2025},
  publisher = {Springer Nature Switzerland},
  series = {Lecture Notes in Computer Science},
  volume = {15965},
  pages = {89--99},
  doi = {10.1007/978-3-032-04978-0_9},
  url = {https://doi.org/10.1007/978-3-032-04978-0_9}
}

@inproceedings{eisenschlos2024selectively,
  title = {Selectively Answering Visual Questions},
  author = {Eisenschlos, Julian and Maina, Hern{\'a}n and Ivetta, Guido and Benotti, Luciana},
  booktitle = {Findings of the Association for Computational Linguistics: ACL 2024},
  year = {2024},
  address = {Bangkok, Thailand},
  publisher = {Association for Computational Linguistics},
  pages = {4219--4229},
  doi = {10.18653/v1/2024.findings-acl.250},
  url = {https://aclanthology.org/2024.findings-acl.250/}
}

@misc{raghu2026dontblink,
  title = {Don't Blink: Evidence Collapse during Multimodal Reasoning},
  author = {Raghu, Suresh and Pandey, Satwik},
  year = {2026},
  url = {https://arxiv.org/abs/2604.04207},
  eprint = {2604.04207},
  archivePrefix = {arXiv}
}

@inproceedings{fang2025dropoutdecoding,
  title = {Enhancing Vision-Language Model Reliability with Uncertainty-Guided Dropout Decoding},
  author = {Fang, Yixiong and Yang, Ziran and Chen, Zhaorun and Zhao, Zhuokai and Zhou, Jiawei},
  booktitle = {Advances in Neural Information Processing Systems},
  year = {2025},
  url = {https://proceedings.neurips.cc/paper_files/paper/2025/hash/db48d94a42706019262ed8304fa658c5-Abstract-Conference.html},
  eprint = {2412.06474},
  archivePrefix = {arXiv}
}

@article{fazli2025caac,
  title={Mitigating Hallucination in Large Vision-Language Models via Adaptive Attention Calibration},
  author={Fazli, Mehrdad and Wei, Bowen and Sari, Ahmet and Zhu, Ziwei},
  journal={arXiv preprint arXiv:2505.21472},
  year={2025},
  url={https://arxiv.org/abs/2505.21472}
}

@inproceedings{he2025mmboundary,
  title={MMBoundary: Advancing MLLM Knowledge Boundary Awareness through Reasoning Step Confidence Calibration},
  author={He, Zhitao and Polisetty, Sandeep and Fan, Zhiyuan and Huang, Yuchen and Wu, Shujin and Fung, Yi R.},
  booktitle={Proceedings of the 63rd Annual Meeting of the Association for Computational Linguistics (Volume 1: Long Papers)},
  pages={16427--16444},
  year={2025},
  doi={10.18653/v1/2025.acl-long.802},
  url={https://aclanthology.org/2025.acl-long.802/}
}

@article{jegham2026visual,
  title = {Visual Reasoning Consistency and Robustness Analysis of Multimodal {LLMs}},
  author = {Jegham, Nidhal and Abdelatti, Marwan F. and Hendawi, Abdeltawab M.},
  journal = {Pattern Recognition},
  year = {2026},
  volume = {172},
  pages = {112765},
  doi = {10.1016/j.patcog.2025.112765},
  url = {https://doi.org/10.1016/j.patcog.2025.112765},
  eprint = {2502.16428},
  archivePrefix = {arXiv}
}

@article{kostumov2024uncertaintyaware,
  title={Uncertainty-Aware Evaluation for Vision-Language Models},
  author={Kostumov, Vasily and Nutfullin, Bulat and Pilipenko, Oleg and Ilyushin, Eugene},
  journal={arXiv preprint arXiv:2402.14418},
  year={2024},
  url={https://arxiv.org/abs/2402.14418}
}

@inproceedings{kriz2025prompt4trust,
  title = {{Prompt4Trust}: A Reinforcement Learning Prompt Augmentation Framework for Clinically-Aligned Confidence Calibration in Multimodal Large Language Models},
  author = {Kriz, Anita and Janes, Elizabeth Laura and Shen, Xing and Arbel, Tal},
  booktitle = {Proceedings of the IEEE/CVF International Conference on Computer Vision Workshops},
  year = {2025},
  pages = {1320--1329},
  url = {https://arxiv.org/abs/2507.09279},
  eprint = {2507.09279},
  archivePrefix = {arXiv}
}

@misc{lau2026umpire,
  title = {Uncertainty Quantification for Multimodal Large Language Models with Incoherence-adjusted Semantic Volume},
  author = {Lau, Gregory Kang Ruey and Dao, Hieu and Lin, Nicole Kan Hui and Low, Bryan Kian Hsiang},
  year = {2026},
  url = {https://arxiv.org/abs/2602.24195},
  eprint = {2602.24195},
  archivePrefix = {arXiv}
}

@inproceedings{liu2024mllmsafety,
  title = {Safety of Multimodal Large Language Models on Images and Texts},
  author = {Liu, Xin and Zhu, Yichen and Lan, Yunshi and Yang, Chao and Qiao, Yu},
  booktitle = {International Joint Conference on Artificial Intelligence},
  year = {2024},
  url = {https://arxiv.org/abs/2402.00357},
  eprint = {2402.00357},
  archivePrefix = {arXiv}
}

@misc{liu2024surveyhallucinationlvlm,
  title = {A Survey on Hallucination in Large Vision-Language Models},
  author = {Liu, Hanchao and Xue, Wenyuan and Chen, Yifei and Chen, Dapeng and Zhao, Xiutian and Wang, Ke and Hou, Liping and Li, Rongjun and Peng, Wei},
  year = {2024},
  url = {https://arxiv.org/abs/2402.00253},
  eprint = {2402.00253},
  archivePrefix = {arXiv}
}

@inproceedings{liu2025uqcalibrationsurvey,
  title = {Uncertainty Quantification and Confidence Calibration in Large Language Models: A Survey},
  author = {Liu, Xiaoou and Chen, Tiejin and Da, Longchao and Chen, Chacha and Lin, Zhen and Wei, Hua},
  booktitle = {Proceedings of the 31st ACM SIGKDD Conference on Knowledge Discovery and Data Mining},
  year = {2025},
  doi = {10.1145/3711896.3736569},
  url = {https://arxiv.org/abs/2503.15850},
  eprint = {2503.15850},
  archivePrefix = {arXiv}
}

@article{miyai2024upd,
  title={Unsolvable Problem Detection: Evaluating Trustworthiness of Vision Language Models},
  author={Miyai, Atsuyuki and Yang, Jingkang and Zhang, Jingyang and Ming, Yifei and Yu, Qing and Irie, Go and Li, Yixuan and Li, Hai and Liu, Ziwei and Aizawa, Kiyoharu},
  journal={arXiv preprint arXiv:2403.20331},
  year={2024},
  url={https://arxiv.org/abs/2403.20331}
}

@misc{nag2025pcsgg,
  title = {Conformal Prediction and {MLLM} aided Uncertainty Quantification in Scene Graph Generation},
  author = {Nag, Sayak and Ghosh, Udita and Bose, Sarosij and Ta, Calvin-Khang and Li, Jiachen and Roy-Chowdhury, Amit K.},
  year = {2025},
  url = {https://arxiv.org/abs/2503.13947},
  eprint = {2503.13947},
  archivePrefix = {arXiv}
}

@misc{padhi2025grounding,
  title = {Calibrating Uncertainty Quantification of Multi-Modal {LLMs} using Grounding},
  author = {Padhi, Trilok and Kaur, Ramneet and Cobb, Adam D. and Acharya, Manoj and Roy, Anirban and Samplawski, Colin and Matejek, Brian and Berenbeim, Alexander M. and Bastian, Nathaniel D. and Jha, Susmit},
  year = {2025},
  url = {https://arxiv.org/abs/2505.03788},
  eprint = {2505.03788},
  archivePrefix = {arXiv}
}

@misc{qiu2026unsaf,
  title = {{UnSAF}: A Self-Assessment Framework of Uncertainty Awareness for Multimodal {LLMs}},
  author = {Qiu, Chen-Meng and Wang, Deng-Bao and Zhang, Min-Ling},
  year = {2026},
  url = {https://openreview.net/forum?id=ohedxNATR9},
  note = {Submitted to ICLR 2026}
}

@article{shorinwa2025surveyuqllm,
  title = {A Survey on Uncertainty Quantification of Large Language Models: Taxonomy, Open Research Challenges, and Future Directions},
  author = {Shorinwa, Ola and Mei, Zhiting and Lidard, Justin and Ren, Allen Z. and Majumdar, Anirudha},
  journal = {ACM Computing Surveys},
  year = {2025},
  doi = {10.1145/3744238},
  url = {https://arxiv.org/abs/2412.05563},
  eprint = {2412.05563},
  archivePrefix = {arXiv}
}

@misc{slyman2025mmb,
  title = {Calibrating {MLLM}-as-a-judge via Multimodal Bayesian Prompt Ensembles},
  author = {Slyman, Eric and Tanjim, Md. Mehrab and Kafle, Kushal and Lee, Stefan},
  year = {2025},
  url = {https://arxiv.org/abs/2509.08777},
  eprint = {2509.08777},
  archivePrefix = {arXiv}
}

@article{srinivasan2024recovl,
  title={Selective ``Selective Prediction'': Reducing Unnecessary Abstention in Vision-Language Reasoning},
  author={Srinivasan, Tejas and Hessel, Jack and Gupta, Tanmay and Lin, Bill Yuchen and Choi, Yejin and Thomason, Jesse and Chandu, Khyathi Raghavi},
  journal={arXiv preprint arXiv:2402.15610},
  year={2024},
  url={https://arxiv.org/abs/2402.15610}
}

@article{tayebati2025cap,
  title={Learning Conformal Abstention Policies for Adaptive Risk Management in Large Language and Vision-Language Models},
  author={Tayebati, Sina and Kumar, Divake and Darabi, Nastaran and Jayasuriya, Dinithi and Krishnan, Ranganath and Amit Ranjan Trivedi},
  journal={arXiv preprint arXiv:2502.06884},
  year={2025},
  url={https://arxiv.org/abs/2502.06884}
}

@article{song2026visualidk,
  title={Delineating Knowledge Boundaries for Honest Large Vision-Language Models},
  author={Song, Junru and Hu, Yimeng and Chen, Yijing and Li, Huining and Li, Qian and Cui, Lizhen and Du, Yuntao},
  journal={arXiv preprint arXiv:2604.26419},
  year={2026},
  url={https://arxiv.org/abs/2604.26419}
}

@misc{kumar2026vlmjudges,
  title = {{VLM} Judges Can Rank but Cannot Score: Task-Dependent Uncertainty in Vision-Language Evaluation},
  author = {Kumar, Divake and Tayebati, Sina and Naik, Devashri and Krishnan, Ranganath and Amit Ranjan Trivedi},
  year = {2026},
  url = {https://arxiv.org/abs/2604.25235},
  eprint = {2604.25235},
  archivePrefix = {arXiv}
}

@article{wang2024mmsap,
  title={MM-SAP: A Comprehensive Benchmark for Assessing Self-Awareness of Multimodal Large Language Models in Perception},
  author={Wang, Yuhao and Liao, Yusheng and Liu, Heyang and Liu, Hongcheng and Yu, Wang and Wang, Yanfeng},
  journal={arXiv preprint arXiv:2401.07529},
  year={2024},
  url={https://arxiv.org/abs/2401.07529}
}

@inproceedings{wang2025tron,
  title = {Sample then Identify: A General Framework for Risk Control and Assessment in Multimodal Large Language Models},
  author = {Wang, Qingni and Geng, Tiantian and Wang, Zhiyuan and Wang, Teng and Fu, Bo and Zheng, Feng},
  booktitle = {International Conference on Learning Representations},
  year = {2025},
  url = {https://proceedings.iclr.cc/paper_files/paper/2025/hash/a1722a6bd1023c026a3d6a570fb3af75-Abstract-Conference.html},
  eprint = {2410.08174},
  archivePrefix = {arXiv}
}

@article{wen2025knowyourlimits,
  title = {Know Your Limits: A Survey of Abstention in Large Language Models},
  author = {Wen, Bingbing and Yao, Jihan and Feng, Shangbin and Xu, Chenjun and Tsvetkov, Yulia and Howe, Bill and Wang, Lucy Lu},
  journal = {Transactions of the Association for Computational Linguistics},
  volume = {13},
  pages = {529--556},
  year = {2025},
  url = {https://aclanthology.org/2025.tacl-1.26/},
  doi = {10.1162/tacl_a_00754}
}

@article{wu2025emocaliber,
  title={EmoCaliber: Advancing Reliable Visual Emotion Comprehension via Confidence Verbalization and Calibration},
  author={Wu, Daiqing and Yang, Dongbao and Ma, Can and Zhou, Yu},
  journal={arXiv preprint arXiv:2512.15528},
  year={2025},
  url={https://arxiv.org/abs/2512.15528}
}

@inproceedings{xia2025surveyuqllm,
  title = {A Survey of Uncertainty Estimation Methods on Large Language Models},
  author = {Xia, Zhiqiu and Xu, Jinxuan and Zhang, Yuqian and Liu, Hang},
  booktitle = {Findings of the Association for Computational Linguistics: ACL 2025},
  year = {2025},
  address = {Vienna, Austria},
  publisher = {Association for Computational Linguistics},
  pages = {21381--21396},
  doi = {10.18653/v1/2025.findings-acl.1101},
  url = {https://aclanthology.org/2025.findings-acl.1101/}
}

@misc{xiao2026vlcalibration,
  title = {{VL-Calibration}: Decoupled Confidence Calibration for Large Vision-Language Models Reasoning},
  author = {Xiao, Wenyi and Xu, Xinchi and Gan, Leilei},
  year = {2026},
  url = {https://arxiv.org/abs/2604.09529},
  eprint = {2604.09529},
  archivePrefix = {arXiv}
}

@article{xuan2025seeing,
  title={Seeing is Believing, but How Much? A Comprehensive Analysis of Verbalized Calibration in Vision-Language Models},
  author={Xuan, Weihao and Zeng, Qingcheng and Qi, Heli and Wang, Junjue and Yokoya, Naoto},
  journal={arXiv preprint arXiv:2505.20236},
  year={2025},
  url={https://arxiv.org/abs/2505.20236}
}

@article{yan2026clintrace,
  title={From Attribution to Abstention: Training-Free Attention-Based Auditing for Clinical Summarization},
  author={Yan, Qianqi and Nguyen, Huy and Srivatsa, Sumana and Bandi, Hari and Wang, Xin Eric and Kenthapadi, Krishnaram},
  journal={arXiv preprint arXiv:2601.16397},
  year={2026},
  url={https://arxiv.org/abs/2601.16397}
}

@inproceedings{yang2025heie,
  title = {{HEIE}: {MLLM}-Based Hierarchical Explainable {AIGC} Image Implausibility Evaluator},
  author = {Yang, Fan and Zhen, Ru and Wang, Jianing and Zhang, Yanhao and Chen, Haoxiang and Lu, Haonan and Zhao, Sicheng and Ding, Guiguang},
  booktitle = {Proceedings of the IEEE/CVF Conference on Computer Vision and Pattern Recognition},
  year = {2025},
  url = {https://arxiv.org/abs/2411.17261},
  eprint = {2411.17261},
  archivePrefix = {arXiv}
}

@article{ye2025scpvlm,
  title={Data-Driven Calibration of Prediction Sets in Large Vision-Language Models Based on Inductive Conformal Prediction},
  author={Ye, Yuanchang and Wen, Weiyan},
  journal={arXiv preprint arXiv:2504.17671},
  year={2025},
  url={https://arxiv.org/abs/2504.17671}
}

@inproceedings{yoon2025videollmsrefuse,
  title={Can Video LLMs Refuse to Answer? Alignment for Answerability in Video Large Language Models},
  author={Yoon, Eunseop and Yoon, Hee Suk and Hasegawa-Johnson, Mark A. and Yoo, Chang D.},
  booktitle={International Conference on Learning Representations},
  year={2025},
  url={https://openreview.net/forum?id=P9VdRQOyqu}
}

@misc{zhang2024vluncertainty,
  title = {{VL-Uncertainty}: Detecting Hallucination in Large Vision-Language Model via Uncertainty Estimation},
  author = {Zhang, Ruiyang and Zhang, Hu and Zheng, Zhedong},
  year = {2024},
  url = {https://arxiv.org/abs/2411.11919},
  eprint = {2411.11919},
  archivePrefix = {arXiv}
}

@misc{zhang2025uncertaintyo,
  title = {Uncertainty-o: One Model-agnostic Framework for Unveiling Uncertainty in Large Multimodal Models},
  author = {Zhang, Ruiyang and Zhang, Hu and Fei, Hao and Zheng, Zhedong},
  year = {2025},
  url = {https://arxiv.org/abs/2506.07575},
  eprint = {2506.07575},
  archivePrefix = {arXiv}
}

@inproceedings{zhao2024firstknow,
  title = {The First to Know: How Token Distributions Reveal Hidden Knowledge in Large Vision-Language Models?},
  author = {Zhao, Qinyu and Xu, Ming and Gupta, Kartik and Asthana, Akshay and Zheng, Liang and Gould, Stephen},
  booktitle = {European Conference on Computer Vision},
  year = {2024},
  url = {https://arxiv.org/abs/2403.09037},
  eprint = {2403.09037},
  archivePrefix = {arXiv}
}

@inproceedings{zhao2025crosspred,
  title = {Can We Predict Performance of Large Models across Vision-Language Tasks?},
  author = {Zhao, Qinyu and Xu, Ming and Gupta, Kartik and Asthana, Akshay and Zheng, Liang and Gould, Stephen},
  booktitle = {Proceedings of the 42nd International Conference on Machine Learning},
  year = {2025},
  url = {https://proceedings.mlr.press/v267/zhao25y.html}
}

@misc{zhi2025srice,
  title = {Seeing and Reasoning with Confidence: Supercharging Multimodal {LLMs} with an Uncertainty-Aware Agentic Framework},
  author = {Zhi, Zhuo and Feng, Chen and Daneshmend, Adam and Orlu, Mine and Demosthenous, Andreas and Yin, Lu and Li, Da and Liu, Ziquan and Rodrigues, Miguel},
  year = {2025},
  url = {https://arxiv.org/abs/2503.08308},
  eprint = {2503.08308},
  archivePrefix = {arXiv}
}

@article{zhu2025mohobench,
  title={MoHoBench: Assessing Honesty of Multimodal Large Language Models via Unanswerable Visual Questions},
  author={Zhu, Yanxu and Duan, Shitong and Zhang, Xiangxu and Sang, Jitao and Zhang, Peng and Lu, Tun and Zhou, Xiao and Yao, Jing and Yi, Xiaoyuan and Xie, Xing},
  journal={arXiv preprint arXiv:2507.21503},
  year={2025},
  url={https://arxiv.org/abs/2507.21503}
}

@article{he2026surveyuncertaintysources,
  title = {Survey of Uncertainty Estimation in {LLMs} - Sources, Methods, Applications, and Challenges},
  author = {He, Jianfeng and Yu, Linlin and Li, Changbin and Yang, Runing and Chen, Fanglan and Li, Kangshuo and Zhang, Min and Lei, Shuo and Zhang, Xuchao and Beigi, Mohammad and Ding, Kaize and Xiao, Bei and Huang, Lifu and Chen, Feng and Jin, Ming and Lu, Chang-Tien},
  journal = {Information Fusion},
  volume = {130},
  pages = {104057},
  year = {2026},
  doi = {10.1016/j.inffus.2025.104057},
  url = {https://www.sciencedirect.com/science/article/pii/S1566253525011194}
}

@article{largeMultimodalEval2025,
  title = {Large Multimodal Models Evaluation: A Survey},
  author = {Zhang, Zicheng and Wang, Junying and Wen, Farong and Guo, Yijin and Zhao, Xiangyu and Fang, Xinyu and Ding, Shengyuan and Jia, Ziheng and Xiao, Jiahao and Shen, Ye and Zheng, Yushuo and Zhu, Xiaorong and Wu, Yalun and Jiao, Ziheng and Sun, Wei and Chen, Zijian and Zhang, Kaiwei and Fu, Kang and Cao, Yuqin and Hu, Ming and Zhou, Yue and Zhou, Xuemei and Cao, Juntai and Zhou, Wei and Cao, Jinyu and Li, Ronghui and Zhou, Donghao and Tian, Yuan and Zhu, Xiangyang and Li, Chunyi and Wu, Haoning and Liu, Xiaohong and He, Junjun and Zhou, Yu and Liu, Hui and Zhang, Lin and Wang, Zesheng and Duan, Huiyu and Zhou, Yingjie and Min, Xiongkuo and Jia, Qi and Zhou, Dongzhan and Zhang, Wenlong and Cao, Jiezhang and Yang, Xue and Yu, Junzhi and Zhai, Guangtao},
  journal = {Science China Information Sciences},
  volume = {68},
  pages = {221301},
  year = {2025},
  doi = {10.1007/s11432-025-4676-4},
  url = {https://link.springer.com/article/10.1007/s11432-025-4676-4}
}

@article{yin2023surveyMLLM,
  title = {A Survey on Multimodal Large Language Models},
  author = {Yin, Shukang and Fu, Chaoyou and Zhao, Sirui and Li, Ke and Sun, Xing and Xu, Tong and Chen, Enhong},
  journal = {National Science Review},
  volume = {11},
  number = {12},
  pages = {nwae403},
  year = {2024},
  doi = {10.1093/nsr/nwae403},
  url = {https://academic.oup.com/nsr/article/11/12/nwae403/7896414},
  eprint = {2306.13549},
  archivePrefix = {arXiv}
}

@misc{madhusudhan2026knowing,
  title = {Knowing When Not to Answer: Evaluating Abstention in Multimodal Reasoning Systems},
  author = {Madhusudhan, Nishanth and Yadav, Vikas and Lacoste, Alexandre},
  year = {2026},
  url = {https://arxiv.org/abs/2604.14799},
  eprint = {2604.14799},
  archivePrefix = {arXiv}
}

@misc{alnazi2026omdbench,
  title = {Omni-Modal Dissonance Benchmark: Systematically Breaking Modality Consensus to Probe Robustness and Calibrated Abstention},
  author = {Al Nazi, Zabir and Dipta, Shubhashis Roy and Parvez, Md Rizwan},
  year = {2026},
  url = {https://arxiv.org/abs/2603.27187},
  eprint = {2603.27187},
  archivePrefix = {arXiv}
}

@misc{ortiz2026abstentionknobs,
  title = {Explicit Abstention Knobs for Predictable Reliability in Video Question Answering},
  author = {Ortiz, Jorge},
  year = {2026},
  url = {https://arxiv.org/abs/2601.00138},
  eprint = {2601.00138},
  archivePrefix = {arXiv}
}

@misc{elyassirad2026conrep,
  title = {CONRep: Uncertainty-Aware Vision-Language Report Drafting Using Conformal Prediction},
  author = {Elyassirad, Danial and Gheiji, Benyamin and Vatanparast, Mahsa and Ahmadzadeh, Amir Mahmoud and Agah, Seyed Amir Asef and Moassefi, Mana and Tavakoli, Meysam and Faghani, Shahriar},
  year = {2026},
  url = {https://arxiv.org/abs/2602.03910},
  eprint = {2602.03910},
  archivePrefix = {arXiv}
}

@inproceedings{azad2026artmaybe,
  title = {The Art of Saying ``Maybe'': A Conformal Lens for Uncertainty Benchmarking in {VLM}s},
  author = {Azad, Asif and Hossain, Mohammad Sadat and Shanto, MD Sadik Hossain and Rahman, M. Saifur and Parvez, Md Rizwan},
  booktitle = {Findings of the Association for Computational Linguistics: EACL 2026},
  year = {2026},
  address = {Rabat, Morocco},
  publisher = {Association for Computational Linguistics},
  pages = {5185--5201},
  doi = {10.18653/v1/2026.findings-eacl.274},
  url = {https://aclanthology.org/2026.findings-eacl.274/}
}

@misc{wang2026vlmuqbench,
  title = {{VLM-UQBench}: A Benchmark for Modality-Specific and Cross-Modality Uncertainties in Vision Language Models},
  author = {Wang, Chenyu and Chen, Tianle and Ahmad, H. M. Sabbir and Batmanghelich, Kayhan and Li, Wenchao},
  year = {2026},
  url = {https://arxiv.org/abs/2602.09214},
  eprint = {2602.09214},
  archivePrefix = {arXiv}
}

@misc{dang2026instinct,
  title = {Instinct vs. Reflection: Unifying Token and Verbalized Confidence in Multimodal Large Models},
  author = {Dang, Yunkai and Jiang, Yifan and Jiang, Yizhu and Chen, Anqi and Li, Wenbin and Gao, Yang},
  year = {2026},
  url = {https://arxiv.org/abs/2604.17274},
  eprint = {2604.17274},
  archivePrefix = {arXiv}
}

@misc{yu2026scoop,
  title = {{SCoOP}: Semantic Consistent Opinion Pooling for Uncertainty Quantification in Multiple Vision-Language Model Systems},
  author = {Yu, Chung-En Johnny and Jalaian, Brian and Bastian, Nathaniel D.},
  year = {2026},
  url = {https://arxiv.org/abs/2603.23853},
  eprint = {2603.23853},
  archivePrefix = {arXiv}
}

@misc{park2026vauq,
  title = {{VAUQ}: Vision-Aware Uncertainty Quantification for {LVLM} Self-Evaluation},
  author = {Park, Seongheon and Oh, Changdae and Choi, Hyeong Kyu and Du, Sean and Li, Sharon},
  year = {2026},
  url = {https://arxiv.org/abs/2602.21054},
  eprint = {2602.21054},
  archivePrefix = {arXiv}
}

@misc{khanmohammadi2026grounded,
  title = {Grounded or Guessing? {LVLM} Confidence Estimation via Blind-Image Contrastive Ranking},
  author = {Khanmohammadi, Reza and Miahi, Erfan and Kaur, Simerjot and Smiley, Charese H. and Brugere, Ivan and Thind, Kundan and Ghassemi, Mohammad M.},
  year = {2026},
  url = {https://arxiv.org/abs/2605.10893},
  eprint = {2605.10893},
  archivePrefix = {arXiv}
}

@inproceedings{popordanoska2026clash,
  title = {{CLASH}: A Benchmark for Cross-Modal Contradiction Detection},
  author = {Popordanoska, Teodora and Karamcheti, Siddharth and Krishna, Ranjay and Fei-Fei, Li},
  booktitle = {Proceedings of the CVPR 2026 Foundation Models for Vision Workshop},
  year = {2026},
  url = {https://openaccess.thecvf.com/content/CVPR2026F/papers/Popordanoska_CLASH_A_Benchmark_for_Cross-Modal_Contradiction_Detection_CVPRF_2026_paper.pdf}
}

@misc{zhang2026modalityconflict,
  title = {When Modalities Conflict: How Unimodal Reasoning Uncertainty Governs Preference Dynamics in {MLLM}s},
  author = {Zhang, Zhuoran and Wang, Tengyue and Gong, Xilin and Shi, Yang and Wang, Haotian and Wang, Di and Hu, Lijie},
  year = {2026},
  url = {https://openreview.net/forum?id=lOQHpdi2o9},
  note = {Submitted to ICLR 2026}
}

@inproceedings{tang2025mupm,
  title = {Analysis of Image-and-Text Uncertainty Propagation in Multimodal Large Language Models with Cardiac MR-Based Applications},
  author = {Tang, Yucheng and Fu, Yunguan and Yi, Weixi and Wang, Yipei and Alexander, Daniel C. and Davies, Rhodri and Hu, Yipeng},
  booktitle = {Medical Image Computing and Computer Assisted Intervention -- MICCAI 2025},
  year = {2025},
  publisher = {Springer Nature Switzerland},
  series = {Lecture Notes in Computer Science},
  volume = {15963},
  pages = {36--45},
  doi = {10.1007/978-3-032-04965-0_4},
  url = {https://doi.org/10.1007/978-3-032-04965-0_4}
}

@misc{chen2026multimodalhallucinationsurvey,
  title = {A Survey of Multimodal Hallucination Evaluation and Detection},
  author = {Chen, Zhiyuan and Min, Yuecong and Zhang, Jie and Yan, Bei and Wang, Jiahao and Wang, Xiaozhen and Shan, Shiguang},
  year = {2026},
  url = {https://arxiv.org/abs/2507.19024},
  eprint = {2507.19024},
  archivePrefix = {arXiv}
}

@inproceedings{bhattacharya2025festa,
  title = {{FESTA}: Functionally Equivalent Sampling for Trust Assessment of Multimodal {LLM}s},
  author = {Bhattacharya, Debarpan and Kulkarni, Apoorva and Ganapathy, Sriram},
  booktitle = {Findings of the Association for Computational Linguistics: EMNLP 2025},
  pages = {12277--12295},
  year = {2025},
  publisher = {Association for Computational Linguistics},
  address = {Suzhou, China},
  doi = {10.18653/v1/2025.findings-emnlp.657},
  url = {https://aclanthology.org/2025.findings-emnlp.657/}
}

@article{tran2026ravqa,
  title = {Knowing When to Answer: Adaptive Confidence Refinement for Reliable Audio-Visual Question Answering},
  author = {Tran, Dinh Phu and Jeong, Jihoon and Wazir, Saad and Kim, Seongah and Do, Thao and Subakan, Cem and Kim, Daeyoung},
  journal = {arXiv preprint arXiv:2602.04924},
  year = {2026},
  url = {https://arxiv.org/abs/2602.04924}
}

@article{aallm2026walking,
  title = {Walking Through Uncertainty: An Empirical Study of Uncertainty Estimation for Audio-Aware Large Language Models},
  author = {Kuan, Chun-Yi and Huang, Wei-Ping and Lee, Hung-yi},
  journal = {arXiv preprint arXiv:2604.25591},
  year = {2026},
  url = {https://arxiv.org/abs/2604.25591}
}

@article{abstaineqa2025robots,
  title = {When Robots Should Say {I} Don't Know: Benchmarking Abstention in Embodied Question Answering},
  author = {Wu, Tao and Zhou, Chuhao and Zhao, Guangyu and Cao, Haozhi and Pu, Yewen and Yang, Jianfei},
  journal = {arXiv preprint arXiv:2512.04597},
  year = {2025},
  url = {https://arxiv.org/abs/2512.04597}
}

@article{grc2026ocr,
  title = {From Plausibility to Verifiability: Risk-Controlled Generative {OCR} with Vision-Language Models},
  author = {Gong, Weile and Zuo, Yiping and Lu, Zijian and He, Xin and Fan, Weibei and Dai, Chen},
  journal = {arXiv preprint arXiv:2603.19790},
  year = {2026},
  url = {https://arxiv.org/abs/2603.19790}
}

@article{ocrprobe2025abstain,
  title = {Reading Between the Lines: Abstaining from {VLM}-Generated {OCR} Errors via Latent Representation Probes},
  author = {Yao, Jihan and Kulshrestha, Achin and Rauschmayr, Nathalie and Roberts, Reed and Zhu, Banghua and Tsvetkov, Yulia and Tombari, Federico},
  journal = {arXiv preprint arXiv:2511.19806},
  year = {2025},
  url = {https://arxiv.org/abs/2511.19806}
}

@inproceedings{radflag2025,
  title = {{RadFlag}: A Black-Box Hallucination Detection Method for Medical Vision Language Models},
  author = {Zhang, Serena and Sambara, Sraavya and Banerjee, Oishi and Acosta, Julian N. and Fahrner, L. John and Rajpurkar, Pranav},
  booktitle = {Proceedings of the 4th Machine Learning for Health Symposium},
  series = {Proceedings of Machine Learning Research},
  volume = {259},
  pages = {1087--1103},
  year = {2025},
  publisher = {PMLR},
  url = {https://proceedings.mlr.press/v259/zhang25c.html}
}

@inproceedings{jian2025clearvqa,
  title = {Teaching Vision-Language Models to Ask: Resolving Ambiguity in Visual Questions},
  author = {Jian, Pu and Yu, Donglei and Yang, Wen and Ren, Shuo and Zhang, Jiajun},
  booktitle = {Proceedings of the 63rd Annual Meeting of the Association for Computational Linguistics (Volume 1: Long Papers)},
  pages = {3619--3638},
  year = {2025},
  publisher = {Association for Computational Linguistics},
  address = {Vienna, Austria},
  doi = {10.18653/v1/2025.acl-long.182},
  url = {https://aclanthology.org/2025.acl-long.182/}
}

@inproceedings{liu2024rightthisway,
  title = {Right This Way: Can {VLM}s Guide Us to See More to Answer Questions?},
  author = {Liu, Li and Yang, Diji and Zhong, Sijia and Tholeti, Kalyana Suma Sree and Ding, Lei and Zhang, Yi and Gilpin, Leilani H.},
  booktitle = {Advances in Neural Information Processing Systems},
  year = {2024},
  url = {https://papers.nips.cc/paper_files/paper/2024/hash/efe4e50d492fedc0dfd2959f3320a974-Abstract-Conference.html}
}

@inproceedings{carot2024,
  title = {Towards Calibrated Robust Fine-Tuning of Vision-Language Models},
  author = {Oh, Changdae and Lim, Hyesu and Kim, Mijoo and Han, Dongyoon and Yun, Sangdoo and Choo, Jaegul and Hauptmann, Alexander and Cheng, Zhi-Qi and Song, Kyungwoo},
  booktitle = {Advances in Neural Information Processing Systems},
  year = {2024},
  url = {https://arxiv.org/abs/2311.01723}
}

@inproceedings{alignvqa2026,
  title = {Refine and Align: Confidence Calibration Through Multi-Agent Interaction in {VQA}},
  author = {Pandey, Ayush and Bardhan, Jai and Jain, Ishita and Hebbalaguppe, Ramya S. and Dhanakshirur, Rohan Raju and Vig, Lovekesh},
  booktitle = {Proceedings of the AAAI Conference on Artificial Intelligence},
  year = {2026},
  url = {https://ojs.aaai.org/index.php/AAAI/article/view/41117}
}

@inproceedings{chartqapro2025,
  title = {{ChartQAPro}: A More Diverse and Challenging Benchmark for Chart Question Answering},
  author = {Masry, Ahmed and Islam, Mohammed Saidul and Ahmed, Mahir and Bajaj, Aayush and Kabir, Firoz and Kartha, Aaryaman and Laskar, Md Tahmid Rahman and Rahman, Mizanur and Rahman, Shadikur and Shahmohammadi, Mehrad and Thakkar, Megh and Parvez, Md Rizwan and Hoque, Enamul and Joty, Shafiq},
  booktitle = {Findings of the Association for Computational Linguistics: ACL 2025},
  pages = {19123--19151},
  year = {2025},
  publisher = {Association for Computational Linguistics},
  doi = {10.18653/v1/2025.findings-acl.978},
  url = {https://aclanthology.org/2025.findings-acl.978/}
}

@article{consensusentropy2025,
  title = {Consensus Entropy: Harnessing Multi-{VLM} Agreement for Self-Verifying and Self-Improving {OCR}},
  author = {Zhang, Yulong and Liang, Tianyi and Huang, Xinyue and Cui, Erfei and Guo, Xu and Chu, Pei and Li, Chenhui and Zhang, Ru and Wang, Wenhai and Liu, Gongshen},
  journal = {arXiv preprint arXiv:2504.11101},
  year = {2025},
  url = {https://arxiv.org/abs/2504.11101}
}

@article{mmarag2026,
  title = {Multimodal Adaptive Retrieval Augmented Generation through Internal Representation Learning},
  author = {Du, Ruoshuang and Sun, Xin and Liu, Qiang and Song, Bowen and Chen, Zhongqi and Wang, Weiqiang and Wang, Liang},
  journal = {arXiv preprint arXiv:2603.00511},
  year = {2026},
  url = {https://arxiv.org/abs/2603.00511}
}

@article{avr2026routing,
  title = {Adaptive Vision-Language Model Routing for Computer Use Agents},
  author = {Liu, Xunzhuo and He, Bowei and Liu, Xue and Luo, Andy and Zhang, Haichen and Chen, Huamin},
  journal = {arXiv preprint arXiv:2603.12823},
  year = {2026},
  url = {https://arxiv.org/abs/2603.12823}
}

@article{ask2act2025,
  title = {Grounding Multimodal {LLM}s to Embodied Agents that Ask for Help with Reinforcement Learning},
  author = {Ramrakhya, Ram and Chang, Matthew and Puig, Xavier and Desai, Ruta and Kira, Zsolt and Mottaghi, Roozbeh},
  journal = {arXiv preprint arXiv:2504.00907},
  year = {2025},
  url = {https://arxiv.org/abs/2504.00907}
}

@article{wieczorek2026variational,
  ids = {wieczorek2026varvqa},
  title = {Variational Visual Question Answering for Uncertainty-Aware Selective Prediction},
  author = {Wieczorek, Tobias Jan and Daun, Nathalie and Khan, Mohammad Emtiyaz and Rohrbach, Marcus},
  journal = {Transactions on Machine Learning Research},
  issn = {2835-8856},
  year = {2026},
  url = {https://openreview.net/forum?id=jtnMIbJIso},
  eprint = {2505.09591},
  archivePrefix = {arXiv}
}

\end{document}